\documentclass{article}

\usepackage{microtype}
\usepackage{graphicx}
\usepackage{array,multirow}
\usepackage{subfigure}
\usepackage{booktabs} 

\usepackage{hyperref}

\usepackage[accepted]{icml2020}

\icmltitlerunning{The Benefits of Pairwise Discriminators for Adversarial Training}

\usepackage{amsmath}
\usepackage{amsthm}
\usepackage{amsfonts}       \usepackage{color}
\usepackage{enumitem}
\usepackage{lipsum}
\usepackage{footmisc}
\usepackage{bm}

\DeclareMathOperator*{\argmax}{argmax}
\DeclareMathOperator*{\argmin}{argmin}

\DeclareMathOperator*{\image}{Im}
\newcommand{\dotp}[2]{\big<\,#1\,,\,#2\,\big>}

\newtheorem{theorem}{Theorem}[section]

\newtheorem{proposition}{Proposition}[section]
\newtheorem{definition}{Definition}[section]

\newtheorem{innercustomprop}{Proposition}
\newenvironment{proposition*}[1]{\renewcommand\theinnercustomprop{#1}\innercustomprop}{\endinnercustomprop}
\newcommand{\tdag}{${}^*$}

\begin{document}

\twocolumn[
\icmltitle{The Benefits of Pairwise Discriminators for Adversarial Training}

\icmlsetsymbol{equal}{*}

\begin{icmlauthorlist}
\icmlauthor{Shangyuan Tong}{equal,csail}
\icmlauthor{Timur Garipov}{equal,csail}
\icmlauthor{Tommi S. Jaakkola}{csail}
\end{icmlauthorlist}

\icmlaffiliation{csail}{MIT Computer Science \& Artificial Intelligence Lab, Cambridge, MA, USA}

\icmlcorrespondingauthor{Shangyuan Tong}{sytong@csail.mit.edu}
\icmlcorrespondingauthor{Timur Garipov}{timur@csail.mit.edu}

\vskip 0.3in
]

\printAffiliationsAndNotice{\icmlEqualContribution} 

\begin{abstract}
Adversarial training methods typically align distributions by solving two-player games. However, in most current formulations, even if the generator aligns perfectly with data, a sub-optimal discriminator can still drive the two apart. Absent additional regularization, the instability can manifest itself as a never-ending game. In this paper, we introduce a family of objectives by leveraging pairwise discriminators, and show that only the generator needs to converge. The alignment, if achieved, would be preserved with any discriminator. We provide sufficient conditions for local convergence; characterize the capacity balance that should guide the discriminator and generator choices; and construct examples of minimally sufficient discriminators. Empirically, we illustrate the theory and the effectiveness of our approach on synthetic examples. Moreover, we show that practical methods derived from our approach can better generate higher-resolution images.
\end{abstract}

\section{Introduction}
\label{sec:intro}

The problem of finding a distributional alignment by means of adversarial training has become a core subroutine across learning tasks, from generative adversarial networks (GANs) \cite{goodfellow2014generative} to  domain-invariant training \cite{ganin2016domain, li2018deep}. For instance, in GANs we seek to align samples from the model with real examples (e.g., images). The generative model in GANs is trained by minimizing a discrepancy or divergence measure between the two distributions. This divergence measure is realized by a discriminator trained to separate real examples from those sampled from the model \cite{nowozin2016f}.

\begin{figure}[h]
\centering
    \includegraphics[width=0.99\linewidth]{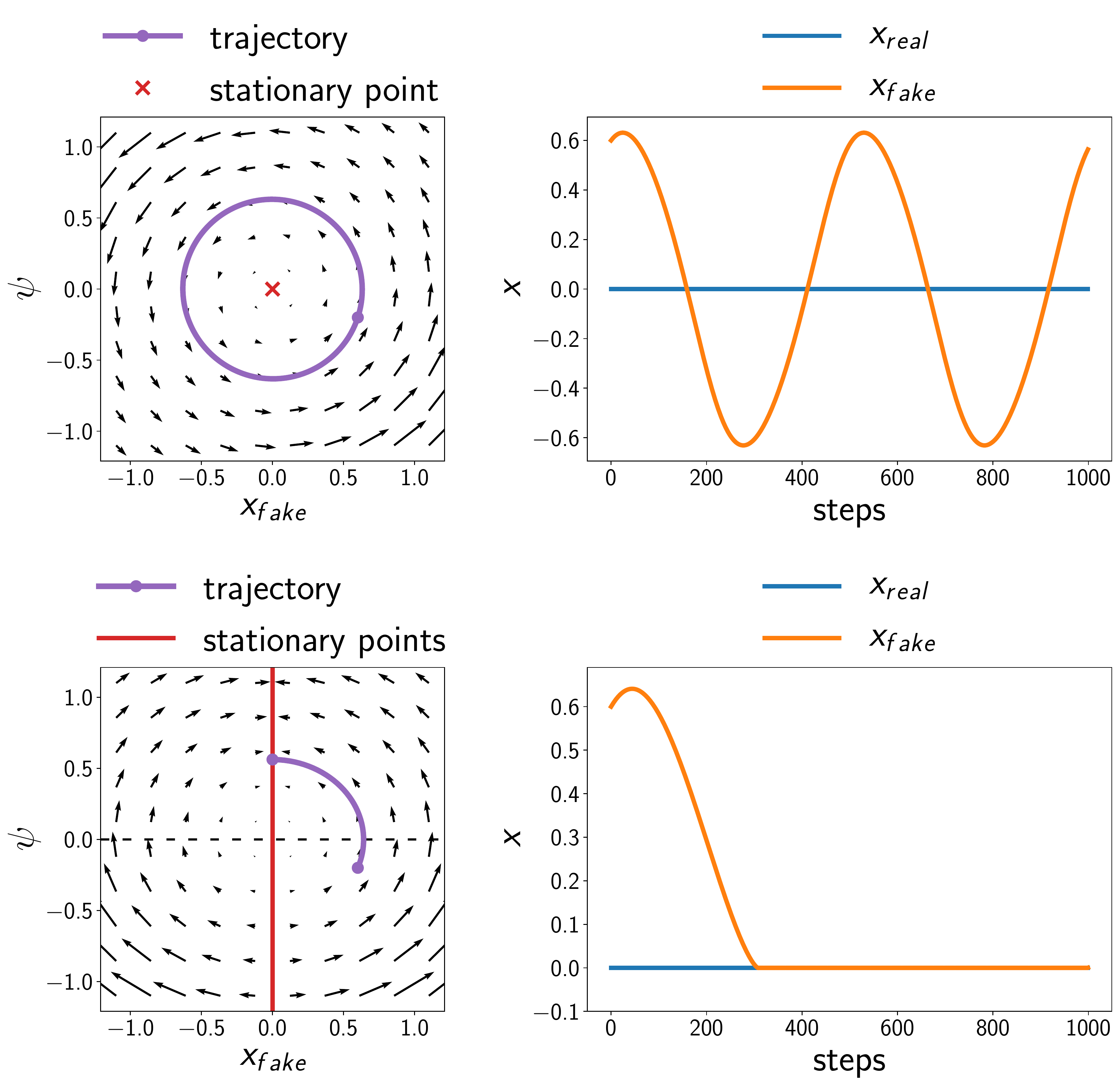}
\caption{Adversarial training demonstrations on toy examples for: \textbf{(top)} unary discriminator with standard objective; \textbf{(bottom)} pairwise discriminator with our objective.
\textbf{Left:} vector field and training trajectories for a toy generator $q(x) = \delta(x - x_\text{fake})$ and a discriminator $D_\psi$ parameterized by a single real number $\psi$.
\textbf{Right:} trajectory of fake and real samples: $x_\text{fake}$ and $x_\text{real} = 0$.
}
\label{fig:toy_example}
\end{figure}

Despite their appeal, GANs are known to be hard to train due to stability issues. Since the estimation is typically setup as two objectives, one for the generator, the other for the discriminator, the desired solution is analogous to a Nash equilibrium of the associated game. Without additional regularization, the dynamics between the two can become unstable \cite{mescheder2018training}, and lead to a never-ending game. While there are multiple reasons for instability, we focus in particular on analyzing and ensuring the stability of alignment around the optimal solution(s). 

The generator sees the training signal, the divergence measure, only through the discriminator. As a result, a generator that aligns perfectly with the target distribution can be thrown off the alignment by a sub-optimal discriminator. In other words, generator can achieve alignment if and only if the discriminator reaches its optimum at the same time.

We illustrate the stability problem and our approach to resolving it with a toy example~\cite{mescheder2018training} in Figure \ref{fig:toy_example}. In this example, the generative model is simply  $q(x) = \delta(x - x_{\text{fake}})$, i.e., concentrated on a single point $x_\text{fake}$, which is the parameter to optimize. The goal is to align it with a real point fixed at $x_{\text{real}} = 0$. The discriminator is a simple classifier $D_\psi(x) = \psi \cdot x$ parameterized by slope $\psi$. The upper-left panel in Figure \ref{fig:toy_example} gives the vector field for an alternating gradient descent training as well as an example trajectory. The training objective is given by the zero-sum game: 
\[
\min\limits_{x_\text{fake}} \max\limits_{\psi}[f(\psi \cdot x_\text{fake}) + f(0)],\;\; f(t) = -\log(1 + e^{-t}).
\]
The upper-right panel shows the time evolution of $x_\text{fake}$ in relation to $x_\text{real}$. Note, in particular, that even when $x_\text{fake}$ reaches the target position $x_\text{real} = 0$ (perfect alignment) the sub-optimal discriminator drives the points apart. 

In this paper, we focus on a different class of discriminators that operate on pairs of samples, trained to identify whether the samples come from the same distribution or not. Utilizing such pairwise discriminators, we identify a family of training objectives which ensures alignment stability even if the discriminator is sub-optimal. 

The panels in the bottom row of Figure \ref{fig:toy_example} illustrate the same example, now with a pairwise discriminator:  $D_\psi(x, y) = \psi \cdot |x - y|^\gamma$, where $\gamma$ is a constant: $\gamma \geq 1$. The lower-left panel again shows the vector field for alternating gradient updates between  $x_\text{fake}$ and $\psi$, resulting from our objective function described in Section \ref{sec:pairgan}. In this case, the alignment $x_\text{fake} = 0$ is a stationary point for \emph{any} discriminator. The time evolution now shows that the alignment is preserved.

We make following contributions:
\begin{enumerate}
    \item In Section \ref{sec:alignment}, we introduce a family of training objectives with pairwise discriminators, that preserve the distribution alignment, if achieved, regardless of the discriminator status.
    \item In Section \ref{sec:pairgan_local}, we show that in our setup, only the generator needs to converge, and we provide sufficient conditions for local convergence.
    \item In Section \ref{sec:pairgan_sufficient}, we introduce the notion of a sufficient discriminator that formalizes the relationship between the capacities of the discriminator and generator. Moreover, we provide constructive examples of minimally sufficient discriminators.
    \item In Section \ref{sec:p1_pn}, we show that our approach and its benefits generalize to aligning multiple distributions.
    \item In Section \ref{sec:experiment}, we show that practical methods derived from our theoretical findings improve the stability and sample quality for generating higher-resolution images. 
\end{enumerate}
The code to reproduce all experiments presented in this paper can be found in \url{https://github.com/ShangyuanTong/PairGAN}.

All proofs can be found in the appendix.

\section{Related work}
\label{sec:related}

\citet{goodfellow2014generative} proposed Generative Adversarial Networks and showed that the associated min-max game can be viewed as minimization of Jensen-Shannon divergence. It was pointed out by \citet{nowozin2016f} that the original GAN objective is a special case of a broader family of min-max objectives corresponding to $f$-divergences. \citet{arjovsky2017wasserstein} showed that game-theoretic setup of GANs can be extended to approximately optimize the Wasserstein distance. \citet{mao2017least} propose LSGAN which uses a least squares objective related to Pearson $\chi^2$ divergence.

\citet{mescheder2017numerics} and \citet{nagarajan2017gradient} proved that GAN training convergences locally for absolutely continuous distributions. However, GANs are commonly used to approximate distributions that lie on low-dimensional manifolds \cite{arjovsky2017towards}. \citet{mescheder2018training} showed that in this case, many training methods do not guarantee local convergence without some additional regularizations, such as instance noise and gradient penalties \cite{roth2017stabilizing}, which enjoy both theoretic guarantees and empirical improvements.

There is a body of work on GANs which utilize pairwise discriminators for improving training dynamics. \citet{jolicoeur-martineau2019relative} has shown that by using a ``relativistic discriminator'', which compares real and fake data, improves both performance and stability. To combat mode collapse, \citet{lin2018pacgan}  proposed to feed a pack of multiple samples from the same distribution to the discriminator rather than a pair of samples from different distributions. Recently, \citet{tsirigotis2019objectives} sought objectives for locally stable GAN training without gradient penalties by using a pairwise discriminator of a specific structure.

We consider an objective which is related to Maximum Mean Discrepancy (MMD) \cite{gretton2007kernel} metric between distributions.  MMD is defined by a positive definite kernel and reaches its minimum only when the distributions are equal. \citet{li2015generative} and \citet{dziugaite2015training} use MMD with RBF-kernels for training deep generative models. In MMD-GAN \cite{li2017mmd, wang2018improving} the kernel function is parameterized by a discriminator-network.

\citet{ganin2016domain} proposed domain-adversarial neural networks (DANN) for domain adaptation. Adversarial training in DANN is used to learn a feature representation such that it can be used to classify the labeled data while keeping the distribution of the representations invariant across the domains. Similarly to GANs, discriminators in DANN are used to estimate a distance between the distributions. \citet{li2018deep} extend this methodology to learn conditional invariant representations for domain generalization.

\section{Background}
\label{sec:notation}

Let $\mathcal{X}$ denote the space of objects (e.g. images). We consider functional spaces $\mathcal{F}(\mathcal{X})$ of real-valued functions $f: \mathcal{X} \to \mathbb{R}$ operating on $\mathcal{X}$. In this paper we consider two particular settings:
\begin{itemize}
    \item $\mathcal{X} = \{x_1, \ldots, x_k\}$ is a finite set, $\mathcal{F}(\mathcal{X}) = \mathbb{R}^k$;
    \item $\mathcal{X} \subset \mathbb{R}^k$ is a compact set, $\mathcal{F}(\mathcal{X})~=~L_2(\mathcal{X})$.
\end{itemize}

In both cases, $\mathcal{F}(\mathcal{X})$ is a vector space with inner product. In our analysis, we build intuition about linear functionals and linear operators on $\mathcal{F}(\mathcal{X})$ treating them as finite-dimensional vectors and matrices. While the space $\mathcal{F}(\mathcal{X}) = \mathbb{R}^k$ provides useful intuition, our results naturally extend to $\mathcal{F}(\mathcal{X}) = L_2(\mathcal{X})$.

\paragraph{Objectives for GANs.} Consider a generative modelling setup where we want to approximate a distribution of ``real'' objects $p(x)$ with a distribution of generated (``fake'') objects $q(x)$. The training in GANs is performed by solving a game between the generator $q(\cdot)$ and a unary discriminator $D(\cdot):~ \mathcal{X}~\to~\mathbb{R}$ which operates on single samples, with the loss functions\footnote{Throughout this paper, we assume that all loss functions are to be minimized, unless stated otherwise.} for the two given by
\begin{subequations}
\label{eq:ugan}
\begin{align}
    \label{eq:ugan_d_loss}
    \mathcal{L}_D &= 
    \mathbb{E}_{p} \Big[ f_1(D(x)) \Big] + 
    \mathbb{E}_{q} \Big[ f_2(D(x)) \Big],
\\
    \label{eq:ugan_g_loss}
    \mathcal{L}_G &= 
    \mathbb{E}_{p} \Big[ g_1(D(x)) \Big] + 
    \mathbb{E}_{q} \Big[ g_2(D(x)) \Big].
\end{align}
\end{subequations}
where $f_1, f_2, g_1, g_2: \mathbb{R} \to \mathbb{R}$ are activation functions applied to the discriminator. The original GAN by \citet{goodfellow2014generative}, which we refer to as the standard GAN (SGAN for short), has $f_1(t) = -g_1(t) = -\log(t)$, $f_2(t) = -g_2(t) = -\log(1-t)$ for the saturating version and $f_1(t) = g_2(t) = -\log(t)$, $f_2(t) = g_1(t) = -\log(1-t)$ for the non-saturating one.

\textbf{Unary discriminators define linear forms.} An expectation
\begin{equation*}
\label{eq:expectation}
\mathbb{E}_{p} \Big[
f(D(x))
\Big] = \int f(D(x)) p(x) \, dx
\end{equation*}
can be viewed as a linear form in the function space: 
\begin{equation*}
    \dotp{a_D^f}{p}
\end{equation*}
where $a_D^f$ and $p$ are the function space vectors corresponding to functions $f(D(\cdot))$ and $p(\cdot)$ respectively.

Using the function space notation we can rewrite the losses \eqref{eq:ugan} as
\begin{subequations}
\label{eq:lin}
\begin{align}
    \label{eq:lin_d_loss}
    \mathcal{L}_D &= 
    \dotp{a_D^{f_1}}{p} + \dotp{a_D^{f_2}}{q},
    \vspace{-2mm}
\\
    \label{eq:lin_g_loss}
    \mathcal{L}_G &= 
    \dotp{a_D^{g_1}}{p} + \dotp{a_D^{g_2}}{q}.
\end{align}
\end{subequations}

\paragraph{Set of probability density functions.}
Note that $p$ and $q$ must define valid density functions. We define $\mathcal{P}(\mathcal{X})$ as the set of probability density functions which belong to $\mathcal{F}(\mathcal{X})$. Formally, we define $\mathcal{P}(\mathcal{X})$ as
\[
\mathcal{P}(\mathcal{X}) = \left\{
    p \in \mathcal{F}(\mathcal{X})
\middle|
    \dotp{p}{e} = 1;~
    p(x) \geq 0~\forall\, x \in \mathcal{X}
\right\},
\]
where $e \in \mathcal{F}{(\mathcal{X})}$ is a function space vector $e(x) \equiv 1$ having the constant value of $1$ on all of its ``positions''. 

\paragraph{Pairwise discriminators.}

The standard setup of GANs can be extended by replacing the unary discriminator
$D(\cdot)$ with a pairwise discriminator $D(\cdot, \cdot): \mathcal{X} \times \mathcal{X} \to \mathbb{R}$ which operates on pairs of samples \cite{li2015generative, jolicoeur-martineau2019relative, jolicoeur2018rfdiv, tsirigotis2019objectives}. In this paper we interpret a pairwise discriminator as a classifier which classifies the pairs of samples into two classes: same distribution pairs and different distribution pairs:
\begin{align*}
    \mbox{same:}&~(x, y) \sim p(x)p(y), \;\;  (x, y) \sim q(x)q(y); \\
    \mbox{different:}&~(x, y) \sim p(x)q(y), \;\;  (x, y) \sim q(x)p(y).
\end{align*}

With a pairwise discriminator, we define a modified game for GANs:
\begin{subequations}
\label{eq:bgan}
\begin{gather}
    \label{eq:bgan_d_loss}
    \begin{split}
        \mathcal{L}_D = 
            \mathbb{E}_{p \times p} \Big[f_1(D(x, y)) \Big]  +
            \mathbb{E}_{q \times q} \Big[f_1(D(x, y)) \Big]
        \\
            + \mathbb{E}_{p \times q} \Big[f_2(D(x, y)) \Big]  +
            \mathbb{E}_{q \times p} \Big[f_2(D(x, y)) \Big] ,
    \end{split}
\\
    \label{eq:bgan_g_loss}
    \begin{split}
        \mathcal{L}_G = 
            \mathbb{E}_{p \times p} \Big[g_1(D(x, y)) \Big]  +
            \mathbb{E}_{q \times q} \Big[g_1(D(x, y)) \Big]
        \\
            +\mathbb{E}_{p \times q} \Big[g_2(D(x, y)) \Big]  +
            \mathbb{E}_{q \times p} \Big[g_2(D(x, y)) \Big] .
    \end{split}
\end{gather}
\end{subequations}

\textbf{Binary discriminators define bi-linear forms.} An expectation:
\begin{equation*}
\mathbb{E}_{p \times q} \Big[
    f(D(x, y))
\Big] \! = \!
\iint \! f(D(x, y)) p(x) q(y) \, dx dy
\end{equation*}
can be viewed as a bi-linear form in the function space:
\begin{equation*}
    \dotp{p}{A_D^f q}
\end{equation*}
where $A_D^f$ denotes a function-space linear operator corresponding to the function $f(D(\cdot, \cdot))$:
\begin{equation*}
    [A_D^f q](x) = \int f(D(x, y)) q(y) \, dy
\end{equation*}

In this paper we consider symmetric discriminators which define
self-adjoint operators:
\[
D(x, y) = D(y, x) ~ \forall\,(x, y) ~ \implies ~
(A_D^f)^T = A_D^f.
\]

Using the bi-linear forms we re-write the losses \eqref{eq:bgan}:
\begin{subequations}
\label{eq:bilin}
\begin{gather}
    \label{eq:bilin_d_loss}
    \begin{split}
        \mathcal{L}_D = 
        \dotp{p}{A_D^{f_1} p} + \dotp{q}{A_D^{f_1} q} \\
        + \dotp{p}{A_D^{f_2} q} + \dotp{q}{A_D^{f_2} p}&,
    \end{split}
\\
    \label{eq:bilin_g_loss}
    \begin{split}
        \mathcal{L}_G = 
        \dotp{p}{A_D^{g_1} p} + \dotp{q}{A_D^{g_1} q} \\
        + \dotp{p}{A_D^{g_2} q} + \dotp{q}{A_D^{g_2} p}&.
    \end{split}
\end{gather}
\end{subequations}

\section{How to preserve the alignment?}
\label{sec:alignment}
\textbf{Unary discriminators destroy the alignment.}
Consider the generator loss for a unary GAN \eqref{eq:lin_g_loss}. 
Suppose that at some moment the generator has been aligned with the target distribution: $q^* = p$. With the subsequent update, $q$ receives the gradient signal $\nabla_q \mathcal{L}_G = a_D^{g_2}$. Below we show that unless $D(x)$ is constant in the support of $p$, the discriminator will drive $q$ away from $p$ and destroy the alignment.

We consider an infinitesimal perturbation $q' = q^* + \varepsilon$. Since $q'$ must be a valid density function, $\varepsilon$ must satisfy:
\[
    \dotp{\varepsilon}{e} = 0, \qquad p(x) + \varepsilon(x) \geq 0 \quad \forall x.
\]

The first-order change of the loss \eqref{eq:lin_g_loss} corresponding to the perturbation $\varepsilon$ is given by:
\[
    \mathcal{L}_G(q^* + \varepsilon, D) - \mathcal{L}_G(q^*, D) \approx \dotp{a_D^{g_2}}{\varepsilon}.
\]
The generator is stationary at $q^*$ iff 
\[
    \dotp{a_D^{g_2}}{\varepsilon} = 0, \quad \forall~\varepsilon: \dotp{\varepsilon}{e} = 0, ~ p(x) + \varepsilon(x) \geq 0 \quad \forall x.
\]
This is only possible when $g_2(D(x))$ is constant in the support of $p$.

This observation implies that the generator can not converge unless the discriminator $D(\cdot)$ converges to the equilibrium position.

\textbf{Pairwise discriminators preserve the alignment.}
We find that there is a family of objectives \eqref{eq:bilin_g_loss} with pairwise discriminators that prevents the discriminator from destroying the alignment, meaning:
\begin{equation}
\label{eq:grad_property}
    \nabla_q \mathcal{L}_G(q, D)\big|_{q = p} = 0, \quad \forall D
\end{equation}
Indeed, in order to satisfy
\begin{equation*}
\nabla_q \mathcal{L}_G(q, D)\big|_{q = p} = 2(A_D^{g_1} + A_D^{g_2})p = 0 \quad \forall\, p
\end{equation*}
it is sufficient to choose $g_2(x) = -g_1(x) \Rightarrow A_D^{g_2} = -A_D^{g_1}$. We define a function $g(x) = g_1(x) = -g_2(x)$ and consider the following instance of the loss \eqref{eq:bilin_g_loss}:
\begin{equation}
\label{eq:g_loss}
    \dotp{p - q}{A_D^g (p - q)}
\end{equation}

\section{PairGAN}
\label{sec:pairgan}
In this section, we first propose PairGAN, a formulation of GANs with the generator loss of the form \eqref{eq:g_loss}. Then, in Section \ref{sec:pairgan_divergencemin}, for specific choices of $f_1, f_2 \text{ and } g$, we provide a theoretical insight similar to that in \cite{goodfellow2014generative} to show that our approach in a specific form also minimizes a meaningful divergence metric. In Section \ref{sec:pairgan_local}, through evaluating the sufficient condition for local convergence, we introduce the notion of sufficient discriminators, which we analyze in details in Section \ref{sec:pairgan_sufficient}.

\paragraph{PairGAN.}
General formulation of PairGAN loss functions are described by a non-zero-sum game:
\begin{subequations}
\label{eq:bgan_v1}
\begin{align}
\label{eq:bgan_v1_d_loss}
    &\begin{aligned}
        L_{\mathcal{D}}(D, q) = 
        \dotp{p}{A_D^{f_1} p} + \dotp{q}{A_D^{f_1} q}& \\
        + \dotp{p}{A_D^{f_2} q} + \dotp{q}{A_D^{f_2} p}&,
    \end{aligned}
\\
\label{eq:bgan_v1_g_loss}
    &L_{\mathcal{G}}(D, q) = \dotp{p - q}{A_D^g (p - q)}.
\end{align}
\end{subequations}

\textbf{PairGAN-Z.}
We also consider a zero-sum game for loss~\eqref{eq:g_loss}. We call the corresponding formulation PairGAN-Z:
\begin{equation}
\label{eq:bgan_v2}
    \min\limits_q \max\limits_D L_{\mathcal{G}}(D, q).
\end{equation}

\subsection{Divergence minimization} \label{sec:pairgan_divergencemin}

Consider the following choice of functions $f_1$, $f_2$ and $g$ for PairGAN \eqref{eq:bgan_v1} and PairGAN-Z \eqref{eq:bgan_v2} where
\[
f_1(t) = -\log(t),~f_2(t) = - \log(1 - t),~g(t) = \log(t).
\]

These loss functions are natural choice for a probabilistic discriminator. In this setup, we interpret the output of a pairwise discriminator $D(\cdot, \cdot)$ as the estimated probability of a pair being sampled from the same distributions. Here, we will show that both non-zero-sum and zero-sum setups minimize meaningful divergence metrics.

Let us define the following mixture distributions:
\begin{subequations}
\label{eq:mix}
\begin{align}
\label{eq:mix_pos}
& M_{p, q}^+(x, y) = \frac{1}{2}\big(
    p(x)p(y) + q(x)q(y)
\big),\\
\label{eq:mix_neg}
& M_{p, q}^-(x, y) = \frac{1}{2}\big(
    p(x)q(y) + q(x)p(y)
\big),\\
\label{eq:mix_all}
& M_{p, q}(x, y) = \frac{1}{2}\left(
    M_{p, q}^+(x, y) + M_{p, q}^-(x, y)
\right).
\end{align}
\end{subequations}

The family of discriminators for PairGAN is defined as:
\[
    \mathcal{D}_{(0, 1)} = \{
        D(\cdot, \cdot)
    ~\vert~
        D(x, y) \in (0, 1) ~ \forall\, (x, y)
    \}.
\]

The generator loss evaluated at the optimal PairGAN discriminator is
\[
\widehat L_\mathcal{G}^1(q) = L_{\mathcal{G}}(D^*(q), q), ~
D^*(q)=\argmin_{D \in \mathcal{D}_{(0, 1)}} L_{\mathcal{D}}(D, q).
\]

For PairGAN-Z, we define another family of probabilistic discriminators whose values are separated from zero:
\[
    \mathcal{D}_{[\varepsilon, 1]} = \{
        D(\cdot, \cdot)
    ~\vert~
        D(x, y) \in [\varepsilon, 1] ~ \forall\, (x, y)
    \},
\]
where $\varepsilon \in (0, 1)$.

Then, the generator loss evaluated at the optimal PairGAN-Z discriminator is
\[
\widehat L_\mathcal{G}^2(q) = L_{\mathcal{G}}(D^*_\varepsilon(q), q), ~
D^*_\varepsilon(q)=\argmax_{D \in \mathcal{D}_{[\varepsilon, 1]}} L_{\mathcal{G}}(D, q).
\]

Now, we can show that with optimal discriminators, these particular choices of PairGAN and PairGAN-Z minimize a symmetrized KL divergence and a total variation distance respectively.
\begin{proposition}
\label{prop:div}
Each of the values $\widehat L_\mathcal{G}^1$ and $\widehat L_\mathcal{G}^2$ is equivalent to a divergence between the distributions $p$ and $q$. Specifically:
\begin{align*}
    & \widehat L_{\mathcal{G}}^1(q) = 4 \cdot \Big( \operatorname{KL}(M_{p, q}^+ \| M_{p, q}) + \operatorname{KL}( M_{p, q} \| M_{p, q}^+) \Big),
\\
    & \widehat L_{\mathcal{G}}^2(q) = - \log(\varepsilon) \cdot \delta_{\text{TV}}(M_{p, q}^+ \| M_{p, q}^-).
\end{align*}
Consequently, for $i \in \{1, 2\}$: 
\[
\widehat L_{\mathcal{G}}^i(q) \geq 0 \quad \forall\, q; \qquad
\widehat L_{\mathcal{G}}^i(q) = 0 ~\Longleftrightarrow~ q = p.
\]
\end{proposition}

\subsection{Local convergence of generator} \label{sec:pairgan_local}
We note that in game \eqref{eq:bgan_v1}, since the generator loss is designed to preserve alignment once achieved, we only require the generator to reach alignment but do not require the discriminator to converge to a specific position. Thus, the goal of our convergence analysis is to identify the set of discriminators which allow the generator to converge. 

Let $D_\psi(\cdot, \cdot)$ and $q(\cdot; \theta)$ be parametric discriminator and generator parameterized by vectors $\psi \in \mathbb{R}^m$ and $\theta \in \mathbb{R}^n$ respectively.

We consider the realizable setup, that is we assume that there exists $\theta^*$ such that: $q(\cdot; \theta^*) = p(\cdot)$. Generally, a parametrization  may permit different instances of parameters to define the same distribution. Hence, we consider a reparametrization manifold \cite{mescheder2018training}:
\[
\label{eq:M_G}
\mathcal{M}_G = \left\{
    \theta 
\middle\vert
    q(\cdot; \theta) = p(\cdot)
\right\}.
\]
In our analysis below, we assume that there is an $\epsilon$-ball $B_\epsilon(\theta^*)$ around $\theta^* \in \mathcal{M}_G$ such that $\mathcal{M}_G \cap B_\epsilon(\theta^*)$ defines a $\mathcal{C}^1$-manifold. We denote the tangent space of the manifold $\mathcal{M}_G$ at $\theta^*$ by $\mathcal{T}_{\theta^*} \mathcal{M}_G$.

Recall from Section \ref{sec:alignment} that $\theta^*$ is a stationary generator for any discriminator $\psi$. Similar to \citet{mescheder2018training}, we analyze the local convergence by examining the eigenvalues of the Hessian of the loss  \eqref{eq:bgan_v1_g_loss} w.r.t $\theta$ at $\theta^*$. We denote this Hessian by $H(\theta^*; \psi) = \nabla^2_{\theta\theta} L_{\mathcal{G}}(D_\psi, q(\cdot; \theta))|_{\theta=\theta^*} $. In Appendix \ref{app:hessian} we show that the Hessian is given by
\begin{multline}
\label{eq:hessian}
H(\theta^*; \psi) = 2 \iint \Big[
    g(D_\psi(x, y)) \\
    \cdot \big(
        [\nabla_\theta q(x; \theta^*)]
        [\nabla_\theta q(y; \theta^*)]^T
    \big)
\Big] dx\,dy.
\end{multline}

The following proposition provides a sufficient condition for local convergence of the generator.
\begin{proposition}
\label{prop:local_conv}
Suppose that $\theta^* \in \mathcal{M}_G$ and a pair $(\psi_0, \theta^*)$ satisfies:
\begin{equation}
\label{eq:PD_param}
    u^T[H(\theta^*; \psi_0)]u > 0 \quad \forall u \notin \mathcal{T}_{\theta^*} \mathcal{M}_G.
\end{equation}
Then, with fixed $\psi = \psi_0$, gradient descent w.r.t. $\theta$ for \eqref{eq:bgan_v1_g_loss} converges to $\mathcal{M}_G$ in a neighborhood of $\theta^*$  provided a small enough learning rate. Moreover, the rate of convergence is at least linear.
\end{proposition}

In Section \ref{sec:pairgan_sufficient}, we clarify what condition \eqref{eq:PD_param} entails. 

Proposition \ref{prop:local_conv} states that a discriminator $\psi_0$ satisfying condition \eqref{eq:PD_param} allows the generator to converge. While, the convergence guarantee is only established for training the generator with a fixed discriminator, this result still holds if we allow $\psi$ to vary within a set. Indeed, from Proposition \ref{prop:local_conv} it follows that $\theta$ converges to $\mathcal{M}_G$, given that $\psi$ remains in the set of the discriminators satisfying \eqref{eq:PD_param}. Note that this set includes all discriminator $\psi$ in a neighborhood of $\psi_0$, since $u^T [H(\theta^*, \psi)]u$ is continuous at $\psi_0$ for any $u$.

\begin{figure}[h]
    \centering
    \includegraphics[width=0.99\linewidth]{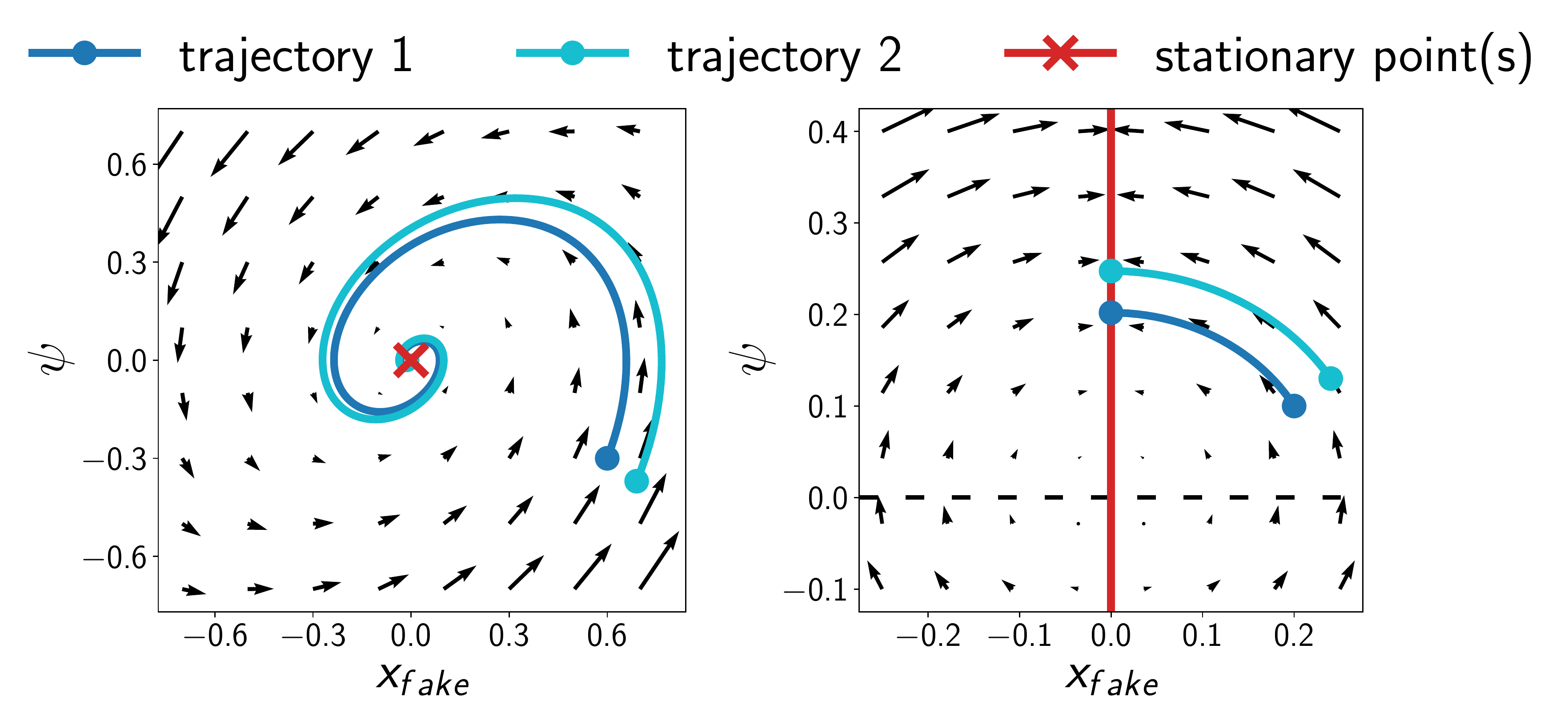}
    \caption{Visualization of convergence points on a toy problem. \textbf{Left:} SGAN with gradient penalty. \textbf{Right:} PairGAN.}
    \label{fig:convergence}
\end{figure}

Figure \ref{fig:convergence} contrasts the convergence properties for GANs with unary discriminators and PairGAN on a toy example identical to that described in Section \ref{sec:intro}. Left panel of Figure~\ref{fig:convergence} shows two trajectories for SGAN with gradient penalties \cite{mescheder2018training}. Both trajectories converge to the only stationary point. In contrast, for PairGAN (Figure \ref{fig:convergence}, right), two trajectories initialized at different points both achieve the alignment $x_\text{fake} = 0$ but converge to different positions of discriminator $\psi$. In this example, the discriminators corresponding to $\psi > 0$ satisfy \eqref{eq:PD_param} and define the gradient vector field pointing towards the line $x_\text{fake}~=~0$. We note that the discriminator updates tend to keep $\psi$ positive. In Section \ref{ssec:global} we extend this observation for PairGAN-Z.

\subsection{Sufficient discriminators} \label{sec:pairgan_sufficient}

To characterize the set of discriminators satisfying condition \eqref{eq:PD_param}, we build intuition from the function space perspective.

We consider a perturbed value of the parameters of the generator $\theta' = \theta^* + u$, where $u \in \mathbb{R}^n$ is an infinitesimal perturbation vector. The corresponding first-order perturbation of the generated distribution $q$ can be expressed via Taylor expansion:
\begin{equation}
\label{eq:eps_u}
    \varepsilon_u(x) = u^T [\nabla_\theta q(x; \theta^*)] \approx q(x; \theta^* + u) - q(x; \theta^*).
\end{equation}

Note that $\varepsilon_u(.)$ is a linear combination of the derivatives w.r.t. to individual parameters $\theta_1, \ldots, \theta_n$:
\[
    \varepsilon_u(x) = \sum\limits_{i=1}^n u_i \alpha_i(x), \qquad \alpha_i(x) = \frac{\partial}{\partial \theta_i} q(x; \theta^*).
\]
Thus, the set of all $\varepsilon_u(x)$ defines a finite-dimensional subspace of the function space. We denote this subspace by $W_q(\theta^*)$:
\[
\label{eq:W_q}
    W_q(\theta^*) := \left\{ \varepsilon(\cdot) \middle| \varepsilon(x) = \sum\limits_{i=1}^n u_i \alpha_i(x),~ u \in \mathbb{R}^n \right\}.
\]
Note that $\dim(W_q(\theta^*)) = n - \dim(\mathcal{M}_G)$, since
\[
    \varepsilon_u(x) \equiv 0 
    \Leftrightarrow
    u^T [\nabla_\theta q(x; \theta^*)] \equiv  0 \Leftrightarrow
    u \in \mathcal{T}_{\theta^*}\mathcal{M}_G.
\]

The expression in equation \eqref{eq:PD_param} can be rewritten in terms of the perturbation $\varepsilon_u$:
\[
u^T [H(\theta^*; \psi)]u = \dotp{\varepsilon_u}{A_{D_\psi}^g \varepsilon_u}.
\]
The following definition gives a function space reformulation of the condition \eqref{eq:PD_param}.
\begin{definition}
\label{def:sufficient}
We say that a self-adjoint operator $A$ is \underline{sufficient} for a parametric generator $q(\cdot; \theta)$ at $\theta^*$ if
\begin{equation}
\label{eq:PD}
\dotp{\varepsilon_u}{A \varepsilon_u} > 0, \quad \forall\, \varepsilon_u \in W_q(\theta^*),~\varepsilon_u \neq 0.
\end{equation}
We say that a discriminator $D$ is \underline{sufficient} for $q(\cdot; \theta)$ at $\theta^*$ if the corresponding operator $A_D^g$ is sufficient for $q(\cdot; \theta)$ at~$\theta^*$.
\end{definition}

This definition essentially means that a discriminator is sufficient for a particular aligned generator if every possible change that this generator can make only result in increasing the generator loss.

Note that for the condition \eqref{eq:PD} to be satisfied it is required that $W_q(\theta^*) \subseteq \image(A)$.

\begin{definition}
We say that an operator $A$ is \underline{minimally} \underline{sufficient} for $q(\cdot; \theta)$ at $\theta^*$ if 
\begin{enumerate}[label=(\roman*)]
    \item $A$ is sufficient for $q(\cdot; \theta)$ at $\theta^*$;
    \item for any sufficient operator $B: \image(A) \subseteq \image(B)$.
\end{enumerate}
\end{definition} 

The following proposition provides constructive examples of minimally sufficient discriminators for any given parametric generator.
\begin{proposition}
\label{prop:operator_example}
Let $g_1(x; \theta)$ and $g_2(x; \theta)$ denote the functions:
\[
g_1(x; \theta) = \nabla_\theta q(x; \theta) \quad
g_2(x; \theta) = \nabla_\theta \log q(x; \theta).
\]
The operators $A^*_1$ and $A^*_2$:
\[
    A^*_i(x, x'; \theta^*) = [g_i(x; \theta^*)]^T [g_i(x; \theta^*)];
\]
are minimally sufficient operators for $q(\cdot; \theta)$ at $\theta^*$.

These operators define the following generator objectives \eqref{eq:bgan_v1_g_loss}:
\begin{equation}
\label{eq:loss_a_star}
L_i^*(\theta) = \Big\|
    \mathbb{E}_{p(x)} \Big[
        g_i(x; \theta)
    \Big]
    - \mathbb{E}_{q(x; \theta)} \Big[
        g_i(x; \theta)
    \Big]
\Big\|^2.
\end{equation}
\end{proposition}

Appendix \ref{app:L_star} provides a detailed discussion on the interpretation of the operators $A_i^*$ and the objectives $L_i^*(\theta)$.

\paragraph{Non-parametric generators and kernels.} Another interpretation of a sufficient discriminator in condition \eqref{eq:PD} is that its corresponding operator needs to be positive definite in the subspace $W_q(\theta^*)$ defined by the generator. This notion naturally extends to positive definite operators defined by a kernel $k(\cdot, \cdot)$. In fact, for a special case of discriminator-operator $A$ defined by a positive definite kernel: $A(x, y) = k(x, y)$, objective \eqref{eq:g_loss} defines Maximum Mean Discrepancy (MMD) \cite{gretton2007kernel} metric which is used as the loss function in MMD-GAN \cite{li2017mmd, binkowski2018demystifying, wang2018improving}.

In this section, we discuss the connections and differences between MMD-GAN and our method. We start by examining the optimization problem \eqref{eq:bgan_v1_g_loss} in the case of non-parametric generator $q$. 

Suppose that the generator $q$ is an arbitrary continuous density function not restricted to a parametric family. Then, the minimization of the loss \eqref{eq:bgan_v1_g_loss} transforms into a constrained optimization problem w.r.t. a function space vector $q$:
\begin{subequations}
\label{eq:q_min}
\begin{align}
\min\limits_q ~& \dotp{p - q}{A (p - q)}, \\
\label{eq:q_e}
\text{s.t.} ~& \dotp{q}{e} = 1, \;\; q(x) \geq 0 \quad \forall\, x.
\end{align}
\end{subequations}
With constraints \eqref{eq:q_e}, $q$ defines a valid density function.

Let $\varepsilon$ be an infinitesimal perturbation of the aligned distribution $q' = q^* + \varepsilon$. In order for $q'$ to remain a valid distribution, we need to restrict the space of possible perturbations $\varepsilon$. We define the set of admissible perturbations as the intersection $W \cap Z$, where
\begin{gather*}
W = \{\varepsilon \,\vert\, \dotp{\varepsilon}{e} = 0\},~
Z = \left \{\varepsilon \,|\, p(x) + \varepsilon(x) \geq 0, \forall x \right\}.
\end{gather*}
Requiring all admissible perturbation to be ``detectable'' by the operator $A$, we obtain a non-parametric version of the condition \eqref{eq:PD}:
\begin{equation}
\label{eq:PD_np}
\dotp{\varepsilon_u}{A \varepsilon_u} > 0, \quad \forall\, \varepsilon_u \in W \cap Z,~\varepsilon_u \neq 0.
\end{equation}
Condition \eqref{eq:PD_np} is a relaxed version of the condition \eqref{eq:PD}, since $W_q(\theta^*) \subseteq W \cap Z$ for any valid parameterization of a distribution $q$. Now, we contrast the difference between the parametric and non-parametric cases.
\begin{itemize}
    \item \textbf{Non-parametric:} the perturbation of $q^*$ is not restricted by a parameterization; thus $A$ is required to be positive definite in the set $W \cap Z$, which is infinite-dimensional.
    \item \textbf{Parametric:} the perturbation of $q^*$ is restricted by a parameterization; thus $A$ is required to be positive definite in finite-dimensional subspace~$W_q(\theta^*)$.
\end{itemize}

With this connection, we can see the key difference between PairGAN and MMD-GAN. MMD-GAN utilizes kernel operators which are positive definite in the functional space $\mathcal{F}(\mathcal{X})$. These operators guarantee that $q^* = p$ is the unique minimizer in \eqref{eq:q_min}. Note that the set of positive definite kernels is a subset of the set of sufficient operators for a given parameteric generator.

\subsection{Towards global convergence of PairGAN-Z}
\label{ssec:global}

Now, we note an interesting property of PairGAN-Z \eqref{eq:bgan_v2}. 

For simplicity, we consider the case of finite $\mathcal{X} = \{x_1, \ldots, x_k\}$. Let $\Delta^k$ denote the probability simplex in $\mathbb{R}^k$. Consider game \eqref{eq:bgan_v2} between a generator $q \in \Delta^k$ and a discriminator-operator $A \in \mathbb{R}^{k \times k}$ given a target distribution $p \in \Delta^k$. Suppose we initialize $q$ and $A$ with
$q_{(0)}$ and $A_{(0)}$ respectively. An iteration of alternating gradient descent is given by:
\begin{subequations}
\label{eq:v2_altgd}
\begin{align}
    &q_{(i + 1)} =  \operatorname{proj}_{\Delta^k} \Big( q_{(i)} - \alpha \cdot 2A_{(i)} [q_{(i)} - p] \Big), \\
    \label{eq:A_sr1_upd}
    &A_{(i + 1)} =  A_{(i)} + \beta \cdot [p - q_{(i + 1)}][p - q_{(i + 1)}]^T,
\end{align}
\end{subequations}
where $\alpha$ and $\beta$ are positive learning rates.

Suppose that at some iteration $A$ is positive definite. Then, each step of the generator decreases the metric~$\dotp{p~-~q}{A(p~-~q)}$ and drives $q$ towards $p$. Furthermore, once $A$ has become positive definite it is guaranteed to remain positive definite after the symmetric rank-$1$ update \eqref{eq:A_sr1_upd}. Thus once $A$ becomes positive definite, $q$ is guaranteed to converge.

We hypothesize that the observed effect opens the possibility to establish global convergence guarantees for PairGAN-Z. Informally, with each gradient update \eqref{eq:A_sr1_upd}, $A$ becomes ``more'' positive definite. Then it remains to prove formally that with updates \eqref{eq:A_sr1_upd} $A$ reaches positive definite state from any starting point $A_{(0)}$. We leave further analysis of this problem for future work.

\section{Aligning multiple distributions}
\label{sec:p1_pn}

In GANs the goal is to align the generated distribution $q$ with a fixed real distribution $p$. In this section, we consider an extended setup for adversarial training,
where our goal is to align multiple distributions $p_1, \ldots, p_N$ together. This setup is a simplified version of the distribution alignment problem arising in domain-invariant training \cite{ganin2016domain, li2018deep}, where adversarial training is used to make the distributions of representations in multiple domains indistinguishable from one another.

We consider the following loss function for  $p_1, \ldots, p_N$:
\begin{multline}
\label{eq:avd_loss_p1_pn}
\mathcal{L}(p_1, \ldots, p_N \vert D) = \sum\limits_{i, j:~i < j}\!\dotp{p_i - p_j}{A_D^g(p_i - p_j)} = \\
=  (N - 1) \sum\limits_{i=1}^N\!\dotp{p_i}{A_D^g p_i} - \sum\limits_{i, j:~i \neq j}\!\dotp{p_i}{A_D^g p_j}
\end{multline}

The next proposition extends property~\eqref{eq:grad_property} for loss \eqref{eq:avd_loss_p1_pn}.
\begin{proposition}
\label{prop:adv_p1_pn}
Suppose that $p_1 = p_2 = \ldots = p_k$ for some $2 \leq k \leq N$. Then $\forall\, i, j \in \{1, \ldots, k\}$:
\[
\nabla_{p_i} \mathcal{L}(p_1, \ldots, p_N \vert D) = \nabla_{p_j} \mathcal{L}(p_1, \ldots, p_N \vert D) \quad \forall D
\]
\end{proposition}

Proposition \ref{prop:adv_p1_pn} states that whenever all distribution in any given subset of $p_1, \ldots p_N$ become mutually aligned they will receive the same gradient. Consequently the alignment within this subset will be preserved.

\begin{figure}[h]
\centering
\includegraphics[width=0.99\linewidth]{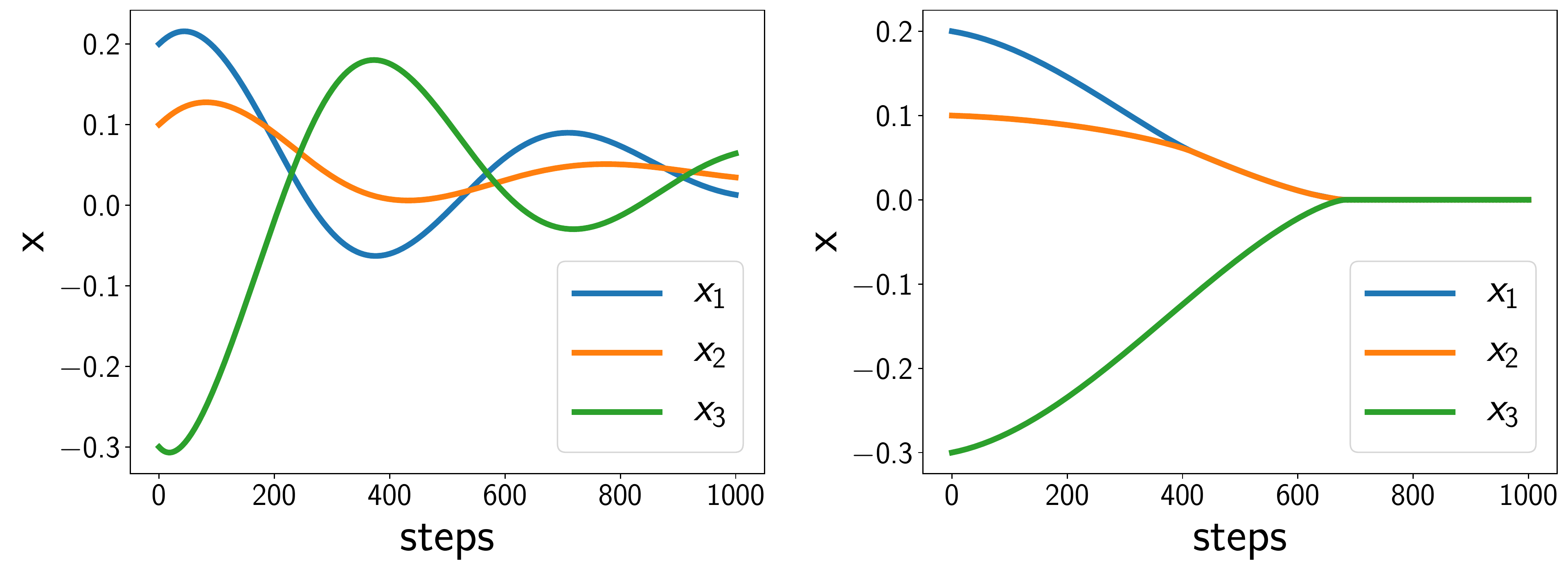}
\caption{Alignment of multiple distributions on a toy example.}
\label{fig:toy_example_mult}
\end{figure}

Figure \ref{fig:toy_example_mult} provides a toy example demonstration for Proposition \ref{prop:adv_p1_pn}. In this example, the goal is to align three distributions $p_i(x) = \delta(x - x_i)$, $i \in \{1, 2, 3\}$.  Both panels on Figure \ref{fig:toy_example_mult} show the trajectories of individual points $x_i$ obtained as result of their interaction with a discriminator (domain-classifier). 
The left panel corresponds to a game with a linear unary discriminator $D(x)$. 
Here, we observe that, when a pair of points becomes aligned, the discriminator can still drive them apart. The right panel of Figure \ref{fig:toy_example_mult} shows the trajectories obtained by using objective \eqref{eq:avd_loss_p1_pn} with a pairwise discriminator $D(x, y) = \psi \cdot |x - y|^{\gamma}$.  We observe that with objective \eqref{eq:avd_loss_p1_pn} the alignment is preserved for any pair of distributions. We provide the detailed specification of the toy example in Appendix \ref{app:toy}.

\section{Experiments}
\label{sec:experiment}

We conducted experiments with a specific form of PairGAN described in Section \ref{sec:pairgan_divergencemin}. Recall, this means that 
\[
    f_1(t) = -\log(t),~f_2(t) = - \log(1 - t),~g(t) = \log(t).
\]

The experiments are set on the CAT dataset \cite{zhang2008cat}, with the same preprocessing setup as \cite{jolicoeur-martineau2019relative}. This is generally a hard problem for generative models because of high-resolution samples (up to 256x256) and small dataset size (about 9k images for 64x64, 6k images for 128x128 and only 2k images for 256x256). The details of our model can be found in Appendix \ref{app:experiment_model}.

We quantitatively evaluate our approach with the Fréchet Inception Distance (FID) \cite{heusel2017gans} (where a lower value generally corresponds to better image quality and sample diversity) on the three choices of resolutions against baselines provided in \cite{jolicoeur-martineau2019relative}. As our specific loss function is a variant of the standard GAN, Table~\ref{table:result} shows our model's performance compared with the baselines that are also variants of standard GAN. These baselines are: standard GAN (SGAN) \cite{goodfellow2014generative}, Relativistic SGAN (RSGAN), Relativistic SGAN with gradient penalty (RSGAN-GP), Relativistic average SGAN (RaSGAN), Relativistic average SGAN with gradient penalty (RaSGAN-GP) \cite{jolicoeur-martineau2019relative}. We find that the baseline for SGAN provided in \cite{jolicoeur-martineau2019relative} uses a numerically unstable implementation of the cross-entropy loss, so we rerun this baseline with the stable implementation. Same as \cite{jolicoeur-martineau2019relative}, we calculate FID at $20$k, $30$k, $\ldots$, $100$k generator steps and report the minimum, maximum, mean and standard deviation of the score values at these $9$ steps. Moreover, we evaluate our method and the fixed SGAN baseline three times and report the average on all four statistics.

\begin{table}[h]
\centering
\caption{Comparsion of minimum, maximum, mean, and standard deviation of FID  calculated at steps 20k, 30k, ..., 100k on different resolutions of CAT dataset \cite{zhang2008cat}. Baseline results denoted with (\tdag) were extracted from \cite{jolicoeur-martineau2019relative}, not independently run in our experiments.}
\vskip 0.15in
 \begin{tabular}{l | c c c c } 
 \toprule
 Loss & Min & Max & Mean & SD \\
 \midrule
 \multicolumn{5}{c}{$64 \times 64$ images}\\
 \midrule
 SGAN & 13.51 & 41.89 & 23.78 & 8.81 \\
 RSGAN\tdag & 19.03 & 42.05 & 32.16 & 7.01 \\
 RaSGAN\tdag & 15.38 & 33.11 & 20.53 & 5.68 \\
 RSGAN-GP\tdag & 16.41 & 22.34 & 18.20 & 1.82 \\
 RaSGAN-GP\tdag & 17.32 & 22 & 19.58 & \textbf{1.81} \\
 \hline
 \textbf{PairGAN (ours)} & \textbf{12.66} & \textbf{20.90} & \textbf{16.38} & 2.23 \\
 \midrule
 \multicolumn{5}{c}{$128 \times 128$ images}\\
 \midrule
 SGAN & 27.35 & 57.76 & 40.17 & 9.34 \\
 RaSGAN\tdag & 21.05 & 39.65 & 28.53 & 6.52 \\
 \hline
 \textbf{PairGAN (ours)} & \textbf{17.30} & \textbf{29.32} & \textbf{21.92} & \textbf{3.76} \\
 \midrule
 \multicolumn{5}{c}{$256 \times 256$ images}\\
 \midrule
 SGAN & 69.64 & 344.55 & 208.99 & 104.08 \\
 RaSGAN\tdag & \textbf{32.11} & 102.76 & 56.64 & 21.03 \\
 \hline
 \textbf{PairGAN (ours)} & 35.35 & \textbf{64.77} & \textbf{45.21} & \textbf{9.49} \\
 \bottomrule
 \end{tabular}
\label{table:result}
\end{table}

From Table \ref{table:result}, we observe that PairGAN improves both performance and stability on higher resolution images. Overall, our method outperforms the baselines in all categories except for RaSGAN-GP (64x64) in standard deviation and RaSGAN (256x256) in minimum.

Appendix \ref{app:experiment_further} provides extended comparison with other baselines, including LSGAN \cite{mao2017least}, HingeGAN \cite{miyato2018spectral}, WGAN-GP \cite{gulrajani2017improved}, and their variants. On 64x64 resolution, PairGAN demonstrates comparable performance with the best baseline (Relativistic average LSGAN) in all four categories. In higher resolution settings (128x128, 256x256), our model achieves the best FID across maximum, mean and standard deviation and its minimum FID is comparable with the best model for that resolution.

We also find our approach to be consistent across multiple runs with small deviations of the four metrics. Further discussion and the full table with the standard deviations of the scores for our model over repeated trials can be found in Appendix \ref{app:experiment_further}. 

We include image samples generated by PairGAN in Appendix \ref{app:experiment_examples}.

\section{Conclusion and future work}
\label{sec:conclusion}

We introduced PairGAN, a formulation of adversarial training where the training dynamics does not suffer from the instability of the alignment. Our theoretical results constitute first steps in understanding convergence guarantees for PairGAN. Interestingly, in our setup, one can formalize the balance of power between the discriminator and the generator with the notion of sufficient discriminators, which is not present in the standard formulation of GANs. 

Directions for future work include further theoretical understanding of convergence guarantees and properties of sufficient discriminators. Throughout our analysis, PairGAN enjoys flexibility which permits the use of different loss functions and model architectures. More extensive experiments with different design choices are necessary to understand the general improvements that PairGAN can bring.

\clearpage

\section*{Acknowledgements}
This work was partially supported by the MIT-IBM collaboration on adversarial learning.

\bibliography{references}
\bibliographystyle{icml2020}

\clearpage
\appendix

\section{Proof of Proposition \ref{prop:div}}
\label{app:div_proof}

Recall that we consider particular instances of the games \eqref{eq:bgan_v1} and \eqref{eq:bgan_v2} corresponding to:
\[
f_1(t) = -\log(t),~f_2(t) = - \log(1 - t),~g(t) = \log(t).
\]

\paragraph{Proof for PairGAN.} First, we expand the expression for the discriminator loss \eqref{eq:bgan_v1_d_loss} in PairGAN:
\begin{flalign*}
L_\mathcal{D}(D, q) =  & + \mathbb{E}_{p(x)p(y)}        \left[ 
        -\log D(x, y)
    \right]\\
    & + \mathbb{E}_{q(x)q(y)} \left[ 
        - \log D(x, y)
    \right]\\
    & + \mathbb{E}_{p(x)q(y)} \left[ 
        - \log (1 - D(x, y))
    \right]\\
    & + \mathbb{E}_{q(x)p(y)} \left[ 
        - \log (1 - D(x, y))
    \right].&
\end{flalign*}
We expand all expectations as the integrals and obtain:
\begin{flalign*}
& L_\mathcal{D} = \iint \Big[ 
        - \log D(x, y) \Big(p(x)p(y) + q(x)q(y)\Big) \\
        & \quad - \log(1 - D(x, y)) \Big(p(x)q(y) + q(x)p(y)\Big)
    \Big]\,dx\,dy.
\end{flalign*}
We minimize the integral by minimizing the expression inside the integral w.r.t $D(x, y)$ point-wise. Solving for the optimal $D(x, y) \in (0, 1)$, we obtain:
\[
    D^*(x, y)\!=\!\frac{p(x)p(y) + q(x)q(y)}{p(x)p(y) +q(x)q(y) + p(x)q(y) + q(x)p(y)}.
\]
We rewrite this expression as the function of the mixture distributions \eqref{eq:mix}
\[
    D^*(x, y) = \frac{M^+_{p, q}(x, y)}{2 \cdot M_{p, q}(x, y)}.
\]

Next, we substitute $D^*$ to the generator loss \eqref{eq:bgan_v1_g_loss}:
\begin{flalign*}
L_\mathcal{G}(D^*, q) =  & + \mathbb{E}_{p(x)p(y)}        \left[ 
        \log D^*(x, y)
    \right]\\
    & + \mathbb{E}_{q(x)q(y)} \left[ 
        \log D^*(x, y)
    \right]\\
    & + \mathbb{E}_{p(x)q(y)} \left[ 
        -\log D^*(x, y)
    \right]\\
    & + \mathbb{E}_{q(x)p(y)} \left[ 
        - \log D^*(x, y)
    \right].&
\end{flalign*}
We add and substract the terms $ \mathbb{E}_{p(x)p(y)}        \left[ 
    \log D^*(x, y)
\right] + \mathbb{E}_{q(x)q(y)} \left[ 
    \log D^*(x, y)
\right]$ to the expression above, and rewrite it as:
\begin{flalign*}
L_\mathcal{G}(D^*, q) =  &                
    +4\mathbb{E}_{M^+_{p,q}(x, y)} \left[ 
        \log D^*(x, y)
    \right] \\
    & - 4 \mathbb{E}_{M_{p, q}(x, y)} \left[ 
        \log D^*(x, y)
    \right] \\
= & +4\mathbb{E}_{M^+_{p,q}(x, y)} \left[ 
        \log \frac{M^+_{p, q}(x, y)}{M_{p, q}(x, y)} - \log(2)
    \right] \\
    & +4\mathbb{E}_{M_{p,q}(x, y)} \left[ 
        \log \frac{M_{p, q}(x, y)}{M^+_{p, q}(x, y)} + \log(2)
    \right].
\end{flalign*}
After cancelling out the constant $\log(2)$ terms, the two expectations above give KL and reverse-KL divergences between $M_{p, q}$ and $M^+_{p, q}$. Thus, we have shown that
\[
L_{\mathcal{G}}(D^*, q) = 4 \cdot \Big( \operatorname{KL}(M_{p, q}^+ \| M_{p, q}) + \operatorname{KL}( M_{p, q} \| M_{p, q}^+) \Big).
\]

The symmetrized KL-divergence above is non-negative and is equal to zero iff 
\[
    M_{p, q}(x, y) =  M_{p, q}^+(x, y)\quad \forall\, x, y.
\]
We transform the last equation in the following way:
\begin{gather*}
    \frac{1}{2}(M_{p, q}^-(x, y) + M_{p, q}^+(x, y)) =  M_{p, q}^+(x, y) \\
    \Updownarrow \\
    M_{p, q}^-(x, y) = M_{p, q}^+(x, y) \\
    \Updownarrow \\
    p(x)q(y) + q(x)p(y) = p(x)p(y) + q(x)q(y) \\
    \Updownarrow \\
    \big(p(x) - q(x)\big) \cdot \big(p(y) - q(y) \big) = 0
\end{gather*}
The last equation holds true for all $(x, y)$ iff $p(\cdot) = q(\cdot)$.

\paragraph{Proof for PairGAN-Z.} First, we expand the expression for the discriminator loss for PairGAN-Z \eqref{eq:bgan_v2}:
\begin{flalign*}
L_\mathcal{G}(D, q) =  & + \mathbb{E}_{p(x)p(y)}\left[ 
        \log D(x, y)
    \right] \\
    & + \mathbb{E}_{q(x)q(y)} \left[ 
         \log D(x, y)
    \right]\\
    & + \mathbb{E}_{p(x)q(y)} \left[ 
        - \log D(x, y)
    \right] \\
    &+ \mathbb{E}_{q(x)p(y)} \left[ 
        - \log D(x, y)
    \right].&
\end{flalign*}

We expand all expectations as the integrals and obtain:
\begin{flalign*}
& L_\mathcal{G} = \iint \Big[ 
    \Big(p(x)p(y) + q(x)q(y)\\
    & \qquad - p(x)q(y) - q(x)p(y)\Big) \log D(x, y)
\Big]\,dx\,dy. &
\end{flalign*}
We introduce the function $F(x, y)$ as:
\[
    F(x, y) = p(x)p(y) + q(x)q(y) - p(x)q(y) - q(x)p(y),
\]
and re-write the loss as
\[
L_\mathcal{G} = \iint \Big[ F(x, y) \log D(x, y)
\Big]\,dx\,dy.
\]
Recall, that in PairGAN-Z the discriminator aims to maximize $L_\mathcal{G}$. Therefore, our goal is to maximize the expression in the integral pointwise w.r.t. $D(x, y) \in [\varepsilon, 1]$. The optimal discriminator $D^*$ is given by\footnote{We restrict the discriminator output $D(x, y) \geq \varepsilon$, in order for the discriminator loss to be bounded. For $F(x, y) < 0$, an unrestricted discriminator can drive $\log D(x, y)$ to $-\infty$.}:
\[
D^*(x, y) = \left\{
\begin{aligned}
    1, \quad & F(x, y) \geq 0 \\
    \varepsilon, \quad & F(x, y) < 0 \\
\end{aligned}
\right..
\]
The logarithm of $D^*$ can be written as:
\[
\log D^*(x, y) = \log(\varepsilon) \mathbb{I}[F(x, y) < 0].
\]
We substitute $\log D^*$ to the generator loss and obtain:
\[
    L_\mathcal{G}(D, q) = \log(\varepsilon) \iint F(x, y) \mathbb{I}[F(x, y) < 0]\, dx\, dy,
\]
where the integral is exactly the negative total variation distance between $M^+_{p, q}$ and $M^-_{p, q}$. Thus, we have shown that:
\[
    L_{\mathcal{G}}(D^*, q) = - \log(\varepsilon) \cdot \delta_{\text{TV}}(M_{p, q}^+ \| M_{p, q}^-).
\]

Similarly to the case of symmetrized KL-divergence above, the total variation distance is non-negative and equals to zero iff
\[
    M_{p, q}^+ = M_{p, q}^- ~\Longleftrightarrow~ p = q.
\]

\section{Hessian of the generator loss}
\label{app:hessian}

For a parametric generator $q(\cdot; \theta)$, we expand the generator loss \eqref{eq:bgan_v1_g_loss}:
\begin{flalign*}
    &L_{\mathcal{G}}(D, q_\theta) = \\
    & \; + \mathbb{E}_{p(x)p(y)} \left[ 
        g(D(x, y))
    \right]
    + \mathbb{E}_{q(x; \theta)q(y; \theta)} \left[ 
        g(D(x, y))
    \right]\\
    &\; - \mathbb{E}_{p(x)q(y; \theta)} \left[ 
        g(D(x, y))
    \right]
    - \mathbb{E}_{q(x; \theta)p(y)} \left[ 
        g(D(x, y))
    \right].&
\end{flalign*}

Now we compute the gradient, by expanding each expectation to an integral and exchanging the order of differentiation and integration:
\begin{flalign*}
& \nabla_\theta L_{\mathcal{G}}(D, q_\theta) = \iint
    \Bigg(
        \nabla_\theta q(x; \theta) \cdot q(y; \theta) + \\
        & \qquad + q(x; \theta) \cdot \nabla_\theta q(y; \theta) - p(x) \cdot \nabla_\theta q(y; \theta) -\\
        & \qquad - \nabla_\theta q(x; \theta) \cdot p(y)
    \Bigg)g(D(x, y))
\, dx\, dy &.
\end{flalign*}

We compute the Hessian by differentiating the gradient:
\begin{flalign*}
& \nabla_\theta L_{\mathcal{G}}(D, q_\theta) = \iint
    \Bigg[ \Bigg(
        2 [\nabla_{\theta} q(x; \theta)] [\nabla_{\theta} q(y; \theta)]^T + 
        \\
        & \qquad + \nabla^2_{\theta \theta} q(x; \theta) \cdot q(y; \theta) + 
        q(x; \theta) \cdot  \nabla^2_{\theta \theta} q(y; \theta) - \\
        &\qquad - \nabla^2_{\theta \theta} q(x; \theta) \cdot p(y) -
        p(x) \cdot  \nabla^2_{\theta \theta} q(y; \theta)
    \Bigg) \times \\
& \quad \times g(D(x, y))\Bigg] \, dx\, dy &.
\end{flalign*}

Our final step is to substitute $\theta = \theta^*$. Since $q(\cdot; \theta^*) = p(x)$, the terms on the second and third lines of the expression above cancel out and we obtain equation \eqref{eq:hessian}.

\section{Proof of Proposition \ref{prop:local_conv}}
\label{app:local_conv}

This sections provides the proof of the Proposition
\ref{prop:local_conv}. The proof relies on the following result by \cite{mescheder2018training}.

\begin{theorem}[Theorem A.3 of \citet{mescheder2018training}]
\label{thm:fixed-point-theorem-manifold}
Let $F(\alpha, \gamma)$ define a $\mathcal C^1$-mapping that maps some domain $\Omega$ to itself.
Assume that there is a local neighborhood $U$ of $0$ such that
$F(0, \gamma) = (0, \gamma)$ for $\gamma \in U$.
Moreover, assume that all eigenvalues of $J:=\nabla_\alpha F(\alpha, 0)\mid_{\alpha=0}$ have absolute value smaller than $1$.
Then the fixed point iteration defined by $F$ is locally convergent to 
$\mathcal M := \{(0, \gamma) \mid \gamma \in U\}$ with linear convergence rate  in a neighborhood of $(0, 0)$.
Moreover, the convergence rate is $|\lambda_{\mathrm{max}}|$ with
$\lambda_\mathrm{max}$ the eigenvalue of $J$
with largest absolute value.
\end{theorem}

\paragraph{Gradient descent update.} We denote the gradient of the loss $L_\mathcal{G}$ w.r.t. $\theta$ as:
\[
g(\theta; \psi) = \nabla_\theta L_\mathcal{G}(D_\psi, q(\cdot; \theta)).
\]
We consider the update operator corresponding to the gradient descent for $L_\mathcal{G}$ w.r.t. $\theta$:
\begin{equation}
 F_h(\theta; \psi) = \theta - h \cdot g(\theta; \psi),
\end{equation}
where $h > 0$ is the step size (learning rate). To understand the convergence of the gradient descent we examine the eigenvalues of the Jacobian $\nabla_\theta F_h(\theta; \psi)$ at $\theta^*$. We notice that $\nabla_\theta F_h(\theta^*; \psi)$ is given by
\[
\nabla_\theta F_h(\theta^*; \psi) = I - h \cdot H(\theta^*; \psi),
\]
where $H(\theta^*; \psi)$ is the Hessian given by \eqref{eq:hessian}. From \eqref{eq:hessian}, we observe that $H(\theta^*; \psi)$ is a symmetric matrix and thus its eigenvalues are real numbers.

An eigenvalue $\lambda$ of the Jacobian $\nabla_\theta F_h(\theta^*; \psi)$ is given by:
\begin{equation}
\label{eq:F_eigval}
\lambda = 1 - h \cdot \mu,
\end{equation}
where $\mu$ is the corresponding eigenvalue of the Hessian $H(\theta^*; \psi)$

Below we provide the proof for Proposition \ref{prop:local_conv}.
\begin{proposition*}{\ref{prop:local_conv}}
Suppose that $\theta^* \in \mathcal{M}_G$ and a pair $(\psi_0, \theta^*)$ satisfies:
\begin{equation}
\label{eq:PD_proof}
    u^T[H(\theta^*; \psi_0)]u > 0 \quad \forall u \notin \mathcal{T}_{\theta^*} \mathcal{M}_G.
\end{equation}
Then, with fixed $\psi = \psi_0$, gradient descent w.r.t. $\theta$ for \eqref{eq:bgan_v1_g_loss} converges to $\mathcal{M}_G$ in a neighborhood of $\theta^*$  provided a small enough learning rate. Moreover, the rate of convergence is at least linear.
\end{proposition*}
\begin{proof}
Following \citet{mescheder2018training}, in order to apply Theorem \ref{thm:fixed-point-theorem-manifold}, we choose local coordinates $\alpha, \gamma$ for $\theta:~\theta(\alpha, \gamma)$. Without loss of generality (see Remark A.6 of \citet{mescheder2018training}), we can assume that
\begin{gather*}
\theta^* = 0, \quad \mathcal{M}_G = \mathcal{T}_\theta^* \mathcal{M}_G = \{0\}^k \times \mathbb{R}^{n - k}, \\
\theta(\alpha, \gamma) = [\alpha, \gamma]^T, \quad \alpha \in \mathbb{R}^k,~ \gamma \in \mathbb{R}^{n - k}.
\end{gather*}
In the local coordinates a vector $u \notin \mathcal{T}_{\theta^*} \mathcal{M}_G$ has the form $u = (\widetilde u, 0)$, where $\widetilde u \in \mathbb{R}^k$. Let $\widetilde H$ denote the sub-matrix of the Hessian $H(\theta^*(\alpha, \gamma); \psi_0)$ corresponding to the coordinates $\alpha$. Then, condition \eqref{eq:PD_proof} transforms into:
\[
\widetilde u^T \widetilde H \widetilde u > 0 \quad \forall\, \widetilde u,
\]
which implies that $\widetilde H$ has only positive eigenvalues. 

In order to apply Theorem \ref{thm:fixed-point-theorem-manifold}, we have to show that all eigenvalues $\lambda$ of the Jacobian $\nabla_\alpha F_h(\theta(\alpha, \gamma); \psi_0)|_{\alpha = 0}$ have absolute value smaller than $1$. Given that $\widetilde H$ has only positive eigenvalues, the inequality $\lambda < 1$ is guaranteed by equation \eqref{eq:F_eigval}.Then it is sufficient for us to choose learning rate $h$ that guarantees $\lambda > -1$. The inequality:
\[
h < \frac{2}{\widetilde \mu_{\text{max}}},
\]
ensures that $\lambda > -1$.

By Theorem \ref{thm:fixed-point-theorem-manifold} the fixed point iteration for $F_h$ converges to $\mathcal{M}_G$.

\end{proof}

\section{Proof of Proposition \ref{prop:operator_example}}
\label{app:operator_example}
We introduce function space operators:
\begin{gather*}
    \Gamma_1, \Gamma_2: \mathbb{R}^n \to \mathcal{F}(\mathcal X), \\
    \begin{aligned}
        \Gamma_1:& \quad  (\Gamma_1 [u])(x) = [g_1(x; \theta)]^T u = [\nabla_\theta q(x; \theta^*)]^T u,\\
        \Gamma_2:& \quad  (\Gamma_2 [u])(x) = [g_2(x; \theta)]^T u =[\nabla_\theta \log q(x; \theta^*)]^T u.
    \end{aligned}
\end{gather*}

Informally, $\Gamma_1, \Gamma_2$ are matrices of size $|\mathcal{X}| \times n$ where the first dimension can be infinite. Let us describe some properties of $\Gamma_1$ and $\Gamma_2$.
\[
    A^*_1 = \Gamma_1 \Gamma_1^T, \qquad
    A^*_2 = \Gamma_2 \Gamma_2^T,
\]
\[
\nabla_\theta \log q(x; \theta) = \frac{1}{q(x; \theta)} \nabla_\theta q(x; \theta) ~ \implies ~ \Gamma_{2} = D_q \Gamma_{\text{1}},
\]
where $D_q$ is a diagonal operator
\[
D_q(x, y) = I[x = y] \frac{1}{q(x; \theta^*)},
\]
with positive values on diagonal\footnote{$q$ must be positive for $\log q$ to be defined.}.

With $\Gamma_1$ we can represent the function-space perturbation \eqref{eq:eps_u} as 
\[
\varepsilon_u = \Gamma_1 u
\]
and re-write Definition \ref{def:sufficient} as
\begin{equation}
\label{eq:sufficient_func}
\Gamma_1\, u \neq 0 \implies u^T \, \Gamma_1^T \, A \, \Gamma_1 \, u > 0.
\end{equation}

By substituting $A = A^*_1$ in \eqref{eq:sufficient_func} we obtain:
\[
\Gamma_1\, u \neq 0 \implies \| \Gamma_1^T \Gamma_1 \, u\|^2 > 0.
\]
This implication holds since $\operatorname{Ker}(\Gamma_1^T) \perp \operatorname{Im}(\Gamma_1)$.

By substituting $A = A^*_2$ in \eqref{eq:sufficient_func}, we obtain:
\[
\Gamma_1\, u \neq 0 \implies \| \Gamma_1^T D_q \Gamma_1 \, u\|^2 > 0,
\]
or equivalently:
\[
\Gamma_1\, u \neq 0 \implies \Big\| \big(D_q^{\frac{1}{2}}\Gamma_1\big)^T \big(D_q^{\frac{1}{2}} \Gamma_1\big) \, u \Big\|^2 > 0.
\]
This implication holds since $\Gamma_1 u \neq 0 \Rightarrow D_q^{\frac{1}{2}} \Gamma_1 u \neq 0$ and $\operatorname{Ker}(D_q^{\frac{1}{2}} \Gamma_1^T) \perp \operatorname{Im}(D_q^{\frac{1}{2}}\Gamma_1)$.

The minimality of the operators follows from the fact that:
\begin{equation}
\label{eq:A_star_rank}
    \operatorname{rank}(A^*_1) = \operatorname{rank}(A^*_2) = \operatorname{dim}(W_q(\theta^*)).
\end{equation}
Recall, that in Section \ref{sec:pairgan_sufficient} we denoted the components of the gradient $\nabla_\theta q(x; \theta^*)$ as function-space vectors $\alpha_1, \ldots, \alpha_n \in \mathcal{F}(\mathcal{X})$:
\[
    \alpha_i(x) = \frac{\partial}{\partial \theta_i} q(x; \theta^*).
\]
Next, we observe that
\[
    W_q(\theta^*) = \operatorname{span}(\alpha_1, \ldots, \alpha_n),
\]
\[ 
    A_1^* = \sum\limits_{i=1}^n \alpha_i \alpha_i^T, \quad 
    A_2^* = \sum\limits_{i=1}^n (D_q \alpha_i) (D_q \alpha_i)^T.
\]
Equation \eqref{eq:A_star_rank} follows from the above representation for $A_1^*$, $A_2^*$ and $W_q(\theta^*)$.

Now, we derive the loss function \eqref{eq:loss_a_star}. We substitute the discriminator-operator $A^*_i$ into the loss \eqref{eq:bgan_v1_g_loss}:
\begin{flalign*}
    L_i^*(\theta) =& \dotp{p - q}{A_i^* (p - q)} = \dotp{p - q}{\Gamma_i \Gamma_i^T (p - q)} \\
    =& \|\Gamma_i^T(p - q)\|^2  = \|\Gamma_i^T p - \Gamma_i^T q\|^2 \\
    =& \Big\|
        \mathbb{E}_{p(x)} \Big[
            g_i(x; \theta)
        \Big]
        - \mathbb{E}_{q(x; \theta)} \Big[
            g_i(x; \theta)
        \Big]
    \Big\|^2
\end{flalign*}

\subsection{Discussion of Proposition \ref{prop:operator_example}}
\label{app:L_star}

Operators $A_i^*$ correspond to discriminators defined through the gradients of the density/log-density of a parametric generator $q(x;\theta)$. In other words, these examples show that given a parametric generator one can construct a minimally sufficient discriminator using the gradients $\nabla_\theta q(x; \theta)$/$\nabla_\theta \log q(x; \theta)$.

Consider the minimization problem for $L_i^*$
\[
    \min\limits_\theta L_i^*(\theta),
\]
which can be written as
\[
    \min\limits_\theta \dotp{p(\cdot) - q(\,\cdot\,; \theta)}{A_i^*(\,\cdot\,,\,\cdot\,; \theta)  [p(\cdot) - q(\,\cdot\,; \theta)]}.
\]
This optimization problem defines a training procedure for the generator $q(x; \theta)$, where instead of training a discriminator, we utilize the operator $A_i^*(\,\cdot\,,\,\cdot\,; \theta)$ which depends on the generator $q(x; \theta)$ itself.

Below, we consider each of the losses $L_1^*(\theta)$ and $L_2^*(\theta)$ and show that they are connected to particular divergence metrics between the distributions $p(x)$ and $q(x; \theta)$.

\textbf{Interpretation of $L_1^*(\theta)$}

We re-write $L_1^*(\theta)$ as
\[
    L_1^*(\theta) = \Big\| \underbrace{
        \mathbb{E}_{q(x; \theta)} \Big[\nabla_\theta q(x; \theta) \Big]
        - \mathbb{E}_{p(x)} \Big[ \nabla_\theta q(x; \theta) \Big]
    }_{I(\theta)} \Big\|^2,
\]
where the function $I(\theta)$ can be expressed as
\begin{flalign*}
    I(\theta) = & \int \big[q(x; \theta) - p(x) \big] \cdot \nabla_\theta q(x; \theta)\, dx \\
    = & \nabla_\theta \Bigg(
          \underbrace{\frac{1}{2} \int \big[q(x; \theta) - p(x) \big]^2\, dx}_{L_{\text{SQ}}(\theta)}
    \Bigg ).
\end{flalign*}
In the above, expression $L_{\text{SQ}}(\theta)$ is a divergence defined by the square of the function-space distance $\|p - q\|$ between $p$ and $q$.

The loss function $L_1^*(\theta)$ is connected to $L_{\text{SQ}}(\theta)$:

\[
    L_1^*(\theta) = \big\| \nabla_\theta L_{\text{SQ}}(\theta) \big\|^2
\]

\textbf{Interpretation of $L_2^*(\theta)$}

Using the fact that 
\[
    \mathbb{E}_{q(x; \theta)} \Big[
        \nabla_\theta \log q(x; \theta)
    \Big] = 0,
\]
we re-write the loss $L_2^*(\theta)$ as:
\[
L_2^*(\theta) = \Big\|
        \mathbb{E}_{p(x)} \Big[
            \nabla_\theta \log q(x; \theta)
        \Big]
    \Big\|^2.
\]

Next, we consider the KL-divergence
\[
    L_{\text{KL}}(\theta) = \text{KL}(p(x) \| q(x; \theta)) = \mathbb{E}_{p(x)} \Big[ 
        \log \frac{p(x)}{q(x; \theta)}
    \Big].
\]
The gradient of $L_\text{KL}(\theta)$ is given by
\[
    \nabla_\theta L_{\text{KL}}(\theta) = 
    - \mathbb{E}_{p(x)} \Big[ 
        \nabla_\theta \log q(x; \theta)
    \Big].
\]

Similarly to $L_1^*$, $L_2^*$ is connected to the KL-divergence:
\[
    L_2^*(\theta) = \big\| \nabla_\theta L_{\text{KL}}(\theta) \big\|^2.
\]

\textbf{Relation to divergence minimization}

Above, we have show that losses $L_1^*$ and $L_2^*$ are connected to the divergences $L_\text{SQ}$ and $L_\text{KL}$ respectively:
\[
    L_1^*(\theta) = \big\| \nabla_\theta L_{\text{SQ}}(\theta) \big\|^2,
    \quad 
    L_2^*(\theta) = \big\| \nabla_\theta L_{\text{KL}}(\theta) \big\|^2.
\]

Every divergence is non-negative and evaluates to zero iff $\theta = \theta^*\!\!\!: q(\,\cdot\,; \theta^*) = p(\cdot)$. Thus, $\theta^*$ is the unique global minimum of both $L_\text{SQ}$ and $L_\text{KL}$\footnote{We note that minimization of the KL-divergence corresponds to maximum likelihood training of the generative model $q(x; \theta)$.}.

We view the minimization of the losses $L_1^*$ and $L_2^*$ as a relaxation of the divergence minimization problem. Each of $L_i^*$ reaches its minimal value $L_i^*(\hat \theta)=0$ iff $\hat \theta$ is a stationary point of the corresponding divergence. In general, a stationary point of $L_\text{SQ}$/$L_\text{KL}$ is not global optimum ($\hat \theta \neq \theta^*$) since both divergences can be non-convex functions of $\theta$.
However, near $\theta^*$, minimization of $L_i^*$ converges to $\theta^*$. Indeed, we are interested in analyzing sufficient operators as they provide guarantees for local convergence for the generator (see Section \ref{sec:pairgan_local}). Propositions \ref{prop:local_conv} and \ref{prop:operator_example} imply that gradient descent for $L_i^*(\theta)$ is locally convergent to $\theta^*$.

\section{Proof of Proposition \ref{prop:adv_p1_pn}}
\label{app:adv_p1_pn}
The gradient of the loss \eqref{eq:avd_loss_p1_pn} w.r.t. $p_i$ is given
\[
\nabla_{p_i} \mathcal{L}(p_1, \ldots, p_N \vert D) = 2(N - 1) A_D^g p_i - \sum\limits_{\substack{s=1\\s\neq i}}^N 2 A_D^g p_s.
\]
For $i \leq k$ we split the sum into two:
\begin{flalign*}
&\nabla_{p_i} \mathcal{L}(p_1, \ldots, p_N \vert D) = \\
&\quad =
    2(N - 1) A_D^g p_i -
    2 \sum\limits_{\substack{s=1\\s\neq i}}^k  A_D^g p_s - 2\sum\limits_{s=k + 1}^N A_D^g p_s.
\end{flalign*}
Next, we use that for $s \leq k: p_s = p_i$, therefore:
\begin{flalign*}
&\nabla_{p_i} \mathcal{L}(p_1, \ldots, p_N \vert D) = \\
& \quad = 2(N - K) A_D^g p_i - 2\sum\limits_{s=k + 1}^N A_D^g p_s.
\end{flalign*}
Finally, we observe that both terms above take the same value for all $1\leq i \leq k$. This observation concludes the proof.

\section{Toy example}
\label{app:toy}

This section provides a detailed description of the toy examples shown in Section \ref{sec:intro} (Figure~\ref{fig:toy_example}), Section \ref{sec:pairgan_local} (Figure~\ref{fig:convergence}), and Section \ref{sec:p1_pn} (Figure~\ref{fig:toy_example_mult}).

Section \ref{app:toy_p_q} describes the toy setup for GANs and the models (unary and pairwise) used to produce Figure~\ref{fig:toy_example} and Figure~\ref{fig:convergence}. Section \ref{app:toy_p1_p3} describes the toy example for multiple distributions alignment (see Section \ref{sec:p1_pn}) and the models  used to produce Figure~\ref{fig:toy_example_mult}. 

The implementation of the described toy examples is provided in the codebase accompanying the paper.

\subsection{DiracGAN \& DiracPairGAN}
\label{app:toy_p_q}

\citet{mescheder2018training} proposed DiracGAN a toy example of GAN, where both target distribution $p$ and generative model $q$ are defined by delta functions (i.e. each concentrated on a single point):
\[
    p(x) = \delta(x - x_\text{real}) \qquad q(x) = \delta(x - x_\text{fake}).
\]
Here, $x_\text{real} = 0$ is a fixed real example, and $x_\text{fake}$ is a free parameter of the generative model $q$. In this model, the distributions are aligned when $x_\text{fake} = x_\text{real}$.

Below we first consider the adversarial training objective for DiracGAN with a simple parameterization of the discriminator used in \cite{mescheder2018training}. Then we introduce DiracPairGAN a modified formulation of DiracGAN with a pairwise discriminator and generator loss of the form \eqref{eq:g_loss}.

\textbf{DiracGAN \cite{mescheder2018training}}

In DiracGAN, the discriminator is defined as linear function $D_\psi(x) = \psi \cdot x$, parameterized by a single number $\psi$. $D_\psi$ defines a linear classifier which estimates the probability of a given sample $x$ being real/fake: 
\begin{align*}
    &P_D(t = \text{real} \,\vert\, x, \psi) = \sigma(D_\psi(x)),\\
    &P_D(t = \text{fake} \,\vert\, x, \psi) = \sigma(-D_\psi(x)),
\end{align*}
where $t \in \{\text{real}, \text{fake}\}$ is a class label and $\sigma(\cdot)$ is the sigmoid function.

The discriminator is trained by maximizing log-likelihood:
\begin{flalign}
\mathcal{L}(\psi, x_\text{fake}) = & + \log P_D(t = \text{real} \,\vert\, x_\text{real}, \psi) 
\nonumber \\
\label{eq:dirac_gan_disc_log_likelihood}
& + \log P_D(t = \text{fake} \,\vert\, x_\text{fake}, \psi).
\end{flalign}

The generator $x_\text{fake}$ and the discriminator $\psi$ compete in a zero-sum game:
\[
    \min\limits_{x_\text{fake}} \max\limits_{\psi} \mathcal{L}(\psi, x_\text{fake}).
\]

Note, that the first term in \eqref{eq:dirac_gan_disc_log_likelihood} is constant since $x_\text{real} = 0$ is constant. Therefore, $\mathcal{L}$ can be equivalently re-written as:
\[
    \widetilde{\mathcal{L}}(\psi, x_\text{fake}) = -\log(1 + \exp\{\psi \cdot x_\text{fake}\}).
\]

It is easy to see that the alignment $x_\text{fake} = 0$ is not preserved in DiracGAN unless $\psi = 0$. To see that it is enough to check that 
\[
    \psi \neq 0 ~\implies~ \frac{\partial}{\partial x} \widetilde{\mathcal{L}}(\psi, 0) \neq 0.
\]

\textbf{DiracPairGAN}

In DiracPairGAN, we define a symmetric pairwise discriminator $D_\psi(x, y) = \psi \cdot |x - y|^\gamma$ where with a single parameter $\psi$ and a hyperparameter $\gamma \geq 1$.

$D_\psi(x, y)$ denotes a probabilistic classifier which estimates the probability of a given pair of samples $(x, y)$ coming from the same distribution rather than different distributions. 
\begin{align*}
    &P_D(t = \text{same} \,\vert\, x, y, \psi) = \sigma(D_\psi(x, y)),\\
    &P_D(t = \text{diff} \,\vert\, x, y, \psi) = \sigma(-D_\psi(x, y)),
\end{align*}
where $t \in \{\text{same}, \text{diff}\}$ denotes the class label.

The negative log-likelihood loss for the pairwise discriminator is given by
\begin{equation}
\label{eq:dirac_pair_gan_D}
\begin{aligned}
    \mathcal{L}_D(\psi, x_\text{fake}) = & -\log P_D(t = \text{same} \,\vert\, x_\text{real}, x_\text{real}, \psi) \\ 
    & - \log P_D(t = \text{same} \,\vert\, x_\text{fake}, x_\text{fake}, \psi) \\
    & - \log P_D(t = \text{diff} \,\vert\, x_\text{real}, x_\text{fake}, \psi) \\
    & - \log P_D(t = \text{diff} \,\vert\, x_\text{fake}, x_\text{real}, \psi).
\end{aligned}
\end{equation}

We define an instance of PairGAN generator loss \eqref{eq:bgan_v1_g_loss}:
\begin{equation}
\label{eq:dirac_pair_gan_G}
\begin{aligned}
    \mathcal{L}_G(\psi, x_\text{fake}) = & -\log P_D(t = \text{diff} \,\vert\, x_\text{real}, x_\text{real}, \psi) \\ 
    & - \log P_D(t = \text{diff} \,\vert\, x_\text{fake}, x_\text{fake}, \psi) \\
    & + \log P_D(t = \text{diff} \,\vert\, x_\text{real}, x_\text{fake}, \psi) \\
    & + \log P_D(t = \text{diff} \,\vert\, x_\text{fake}, x_\text{real}, \psi).
\end{aligned}
\end{equation}

In general formulation of PairGAN the generator and the discriminator compete in a non-zero sum game:
\begin{gather*}
    \min\limits_\psi \mathcal{L}_D(\psi, x_\text{fake}) \\
    \min\limits_{x_\text{fake}} \mathcal{L}_G(\psi, x_\text{fake}).
\end{gather*}

We note, that our choice of parameterization allows us to re-write the game in a simplified form. Indeed, the first two terms in both \eqref{eq:dirac_pair_gan_D} and \eqref{eq:dirac_pair_gan_G} are constant and all equal to $-\log(\frac{1}{2})$ since $D_\psi(x, x) = 0$. Thus, the only difference in the losses \eqref{eq:dirac_pair_gan_D} and \eqref{eq:dirac_pair_gan_G} is in the signs of the third and the fourth terms. Observing this, we obtain an equivalent zero-sum game:
\[
    \min\limits_{x_\text{fake}} \max\limits_{\psi} \widetilde{\mathcal{L}}(\psi, x_\text{fake}),
\]
where
\[
    \widetilde{\mathcal{L}}(\psi, x_\text{fake}) = - \log(1 + \exp\{\psi \cdot |x_\text{fake}|^\gamma\}).
\]

In DiracPairGAN the alignment is preserved for any $\psi$ since $\widetilde{\mathcal{L}}(\psi, x_\text{fake})$ is a function of absolute value of $x_\text{fake}$ and, consequently,
\[
    \frac{\partial}{\partial x} \widetilde{\mathcal{L}}(\psi, 0) = 0 \quad \forall\,\psi.
\]

\subsection{Multiple distributions}
\label{app:toy_p1_p3}

Below we consider the toy example demonstrating adversarial alignment of multiple distributions (see Section \ref{sec:p1_pn}).

Consider, three delta functions: $p_1, p_2, p_3$:
\[ 
    p_i(x) = \delta(x - x_i),
\]
parameterized by real numbers $x_1$, $x_2$, and $x_3$ respectively. The goal of the toy models described below is to align the three distributions with one another, i.e. reach a situation where $x_1 = x_2 = x_3$.

\textbf{Unary discriminator}

For the three distributions problem we utilize a unary discriminator 
\[
    D_{\bm{\psi}}(x) = [s_1(x, \bm{\psi}), s_2(x, \bm{\psi}), s_3(x, \bm{\psi})],
\]
which defines a 3-class softmax classifier
\[
P_D(t = i \,\vert\, x, \bm{\psi}) = \frac{\exp\{s_i(x, \bm{\psi})\}}{\sum\limits_{j=1}^3\exp\{s_j(x, \bm{\psi})\}},
\]
where $t \in \{1, 2, 3\}$ is a class label and $P_D(t = i \,\vert\, x, \bm{\psi})$ is an estimate of the probability of a given sample $x$ coming from $p_i(\cdot)$.

We define the logits $s_i$ as quadratic parametric functions
\[
s_i(x, \bm{\psi}) = a_i x^2 + b_i x + c_i, \quad i \in \{1, 2, 3\}
\]
with $\bm{\psi}$ defined as a vector of all nine parameters
\[
\bm{\psi} = [a_1, b_1, c_1, a_2, b_2, c_2, a_3, b_3, c_3]^T.
\]

One can think of $D_\psi(x)$ as a linear classifier operating on a non-linear feature representation $\bm{\phi}(x) = [x^2, x, 1]^T$:
\[
    s_i(x, \bm{\psi}) = [a_i, b_i, c_i] [x^2, x, 1]^T.
\]
Note that with the described parameterization, the unary discriminator is powerful enough to represent a zero-error decision boundary for any location of the points $x_1, x_2, x_3$.

Similarly to the examples above, we train discriminator by maximizing log-likelihood
\begin{equation}
    \label{eq:domain_loss}
    \mathcal{L}(\bm{\psi}, x_1, x_2, x_3) = \sum \limits_{i = 1}^3 \log P_D(t = i \,\vert\, x_i, \bm{\psi}),
\end{equation}
and define a zero-sum game between the points $x_1$, $x_2$, $x_3$ and discriminator $\bm{\psi}$:
\[
\min\limits_{x_1, x_2, x_3} \max\limits_{\bm{\psi}} \mathcal{L}(\bm{\psi}, x_1, x_2, x_3).
\]

\textbf{Pairwise discriminator}

Now, we define a multiple distributions model with a pairwise discriminator. Again, we utilize the same pairwise discriminator as in Section \ref{app:toy_p_q}: $D_\psi(x, y) = \psi \cdot |x - y|^\gamma$:
\begin{align*}
    &P_D(t = \text{same} \,\vert\, x, y, \psi) = \sigma(D_\psi(x, y)),\\
    &P_D(t = \text{diff} \,\vert\, x, y, \psi) = \sigma(-D_\psi(x, y)).
\end{align*}

We use the following weighted negative log-likelihood objective for the discriminator: 
\begin{equation}
\label{eq:p1_p3_d_loss}
\begin{aligned}
    & \mathcal{L}_D(\psi, x_1, x_2, x_3) = \\
    & \qquad - 2 \sum\limits_{i = 1}^3 \log P_D(t = \text{same} \,\vert\, x_i, x_i, \psi) \\
    & \qquad - \sum\limits_{i, j: i \neq j} \log P_D(t = \text{diff} \,\vert\, x_i, x_j, \psi),
\end{aligned}
\end{equation}
computed for same distribution pairs $(x_i, x_i)$ and different distributions pairs $(x_i, x_j): i \neq j$. In order to equalize the $3:6$ ratio of the number of  $\{t=\text{same}\}$-pairs to the number of $\{t=\text{diff}\}$-pairs, we virtually augment the set of same distribution pair by using the weights $w_\text{same} = 2$, $w_\text{diff} = 1$.

Next, we define a non-zero sum game between $\psi$ and $x_1, x_2, x_3$:
\begin{gather*}
    \min\limits_\psi \mathcal{L}_D(\psi, x_1, x_2, x_3), \\
    \min\limits_{x_1, x_2, x_3} \mathcal{L}_G(\psi, x_1, x_2, x_3),
\end{gather*}
where the loss for $x_1, x_2, x_3$ is an instance of the adversarial loss \eqref{eq:avd_loss_p1_pn} introduced in Section \ref{sec:p1_pn}:
\begin{equation}
\label{eq:p1_p3_g_loss}
\begin{aligned}
    & \mathcal{L}_G(\psi, x_1, x_2, x_3) = \\
    & \qquad - 2 \sum\limits_{i = 1}^3 \log P_D(t = \text{diff} \,\vert\, x_i, x_i, \psi) \\
    & \qquad + \sum\limits_{i, j: i \neq j} \log P_D(t = \text{diff} \,\vert\, x_i, x_j, \psi).
\end{aligned}
\end{equation}

Since $D_\psi(x, x) = 0$, the first terms in both \eqref{eq:p1_p3_d_loss} and \eqref{eq:p1_p3_g_loss} are constant. Therefore, the considered setup can be reduced to a zero-sum game:
\[
    \min\limits_{x_1, x_2, x_3} \max\limits_{\psi} \widetilde{\mathcal{L}}(\psi, x_1, x_2, x_3),
\]
where the loss $\widetilde{\mathcal{L}}$ is given by
\[
    \widetilde{\mathcal{L}}(\psi, x_1, x_2, x_3) = -\sum\limits_{i, j: i \neq j} \log(1 + \exp\{\psi \cdot |x_i - x_j|^\gamma\}).
\]

\paragraph{Comments on domain-adversarial methods for domain-adaptation.}
The loss \eqref{eq:domain_loss} used in the unary discriminator above is a simplified instance of the domain loss used in domain adversarial neural networks (DANN) \cite{ganin2016domain}. In domain adversarial training notation, $\{p_i(x)\}_{i=1}^3$ represent the distribution of representations in different domains. The domain loss \eqref{eq:domain_loss} is one of the terms in DANN objective. Optimization of the domain loss \eqref{eq:domain_loss} w.r.t. parameters of distribution $\{p_i(x)\}_{i=1}^3$ can be interpreted as minimization of a divergence between the distributions. This regularization mechanism is expected to make the learned representation $x$ invariant across the domains.

Note that this example represents only a part of the adversarial objective used in DANN. The full adversarial training procedure is defined as a three-player game between a feature extractor, a classifier and a domain discriminator. In contrast, here we only focus on one loss term which is responsible for the alignment of the distributions. Moreover, in DANN the distributions $\{p_i(x)\}_{i=1}^3$ are interconnected through the shared parameterization, while in the presented toy model we consider independently parameterized distributions. We believe that understanding the mechanics of the alignment with this toy example is important for the analysis and further development of domain-adversarial methods.

\section{Experiments details}
\label{app:experiment}

\begin{figure*}[t]
    \centering
    \includegraphics[height=13em]{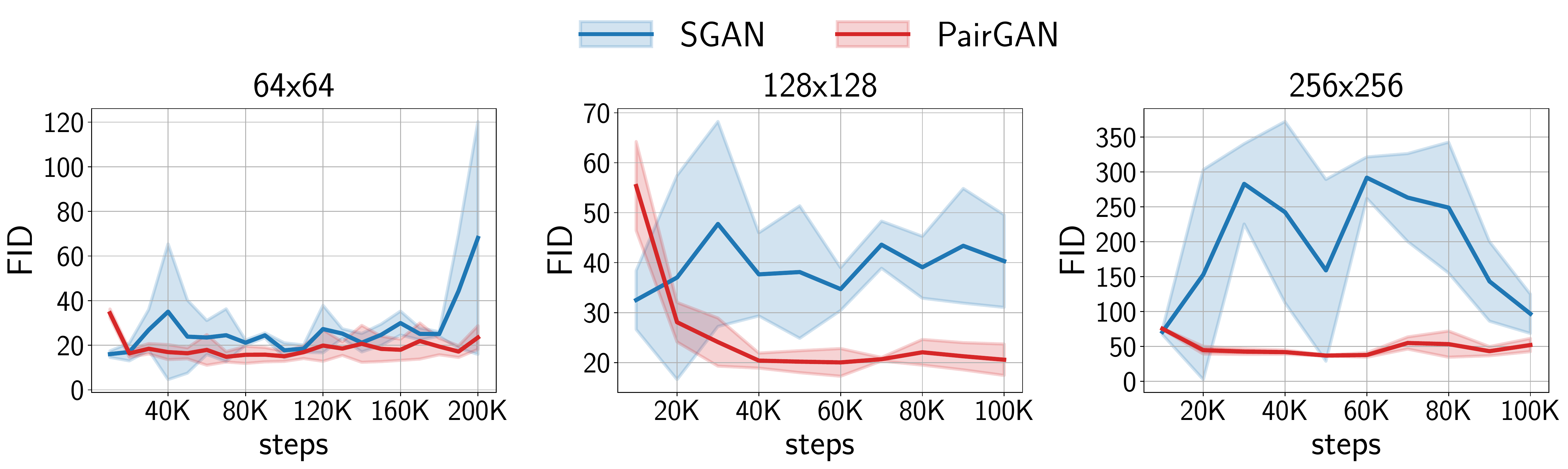}
    \caption{
        FID as a function of training step number for SGAN (blue) and PairGAN (red) on CAT dataset for resolution 64x64, 128x128, and 256x256. Each curve shows the FID statistics at a given step calculated over 3 training runs. The thick line shows mean value and the shaded area represents $\pm 1$ standard deviation range.
    }
    \label{fig:app_stability}
\end{figure*}

\subsection{Model details}
\label{app:experiment_model}
For our PairGAN model, we choose the loss function as described in \ref{sec:pairgan_divergencemin}:
\[
    f_1(t)=-\log(t), f_2(t)=-\log(1-t), g(t)=\log(t).
\]
We utilize the standard DCGAN structure \cite{radford2015unsupervised} with a discriminator architecture modified to represent a symmetric pairwise function. We parameterize the pairwise discriminator as $D(x, y) = D_b(D_u(x)+D_u(y))$ \cite{zaheer2017deep} where $D_u$ is the DCGAN unary discriminator that takes a single image as an input and returns a multi-dimensional output instead of just one-dimensional. $D_b$ is an additional binary network that takes in $D_u$ output and returns a single scalar.

Admittedly, the computation of PairGAN objectives requires sampling of pairs, which increases complexity from $\mathcal{O}(n)$ to $\mathcal{O}(n^2)$ compared to unary discriminator objectives. To address this issue, we find that an averaging approach similar to that in \cite{jolicoeur-martineau2019relative} works well in practice. The modified objective functions then are:
\begin{subequations}
    \begin{flalign}
        \mathcal{L}_D =\,
            &-\mathbb{E}_{p} \Big[\log(\overline{D}(x, p))\Big] 
            -\mathbb{E}_{q} \Big[\log(\overline{D}(x, q))\Big] \nonumber \\
            &- 2\cdot\mathbb{E}_{p} \Big[\log(1-\overline{D}(x, q))\Big],
            \label{eq:emprical-bgan_d_loss} \\
        \mathcal{L}_G =\,& 
            \mathbb{E}_{q} \Big[\log(\overline{D}(x, q))\Big]
            - 2\cdot\mathbb{E}_{p} \Big[\log(\overline{D}(x, q))\Big],
            \label{eq:emprical-bgan_g_loss}
    \end{flalign}
\end{subequations}
where
\begin{align*}
    \overline{D}(x, p) &= D_b\Big(D_u(x)+\mathbb{E}_{p} \Big[D_u(y)\Big]\Big) , \\
    \overline{D}(x, q) &= D_b\Big(D_u(x)+\mathbb{E}_{q} \Big[D_u(y)\Big]\Big) .
\end{align*}

In our experiments, we observe that with the averaged loss function we can implement pairwise discriminator with very little computational overhead compared to unary discriminator used in DCGAN. We find that we can use low-dimensional output layer in the unary network $D_u$. Specifically, we use just 2 dimensions for our experiments. Moreover, for the binary network we use a 1-hidden-layer fully-connected architecture
\[
D_b(x) = \texttt{FC}_{16 \to 1}(\texttt{SELU}(\texttt{FC}_{2 \to 16}(x))),
\]
with SELU activation function \cite{klambauer2017self} and 16 units in the hidden layer.

Similar to the findings in \citet{goodfellow2014generative}, directly optimizing \eqref{eq:emprical-bgan_g_loss} is problematic in the beginning, as the the first term does not provide useful gradients early on. When the generator is underfitted, the discriminator can confidently classify (fake, fake) pairs. If $\overline{D}(x, q) \approx 1$ for $x \sim q$, the loss $\log(\overline{D}(x, q))$ saturates and the gradient for the generator vanishes (since in practice the output of $D$ is parameterized through a sigmoid activation). To address this issue, we start training with a non-saturating loss function and gradually anneal to the desired version in \eqref{eq:emprical-bgan_g_loss}. The annealed loss function for the generator is given by:
\begin{subequations}
\begin{align*}
    \mathcal{L}_G =&+
    \alpha\cdot\mathbb{E}_{q} \Big[-\log(1-\overline{D}(x, q))\Big] \\&+(1-\alpha)\cdot\mathbb{E}_{q} \Big[\log(\overline{D}(x, q))\Big]
    \\&-
    2\cdot\mathbb{E}_{p} \Big[\log(\overline{D}(x, q))\Big]
\end{align*}
the annealing coefficient $\alpha$ changes from $1$ to $0$ as a function of the step counter $i = 1, 2, \ldots$
\begin{equation*}
\alpha_i = \begin{cases}
1,  & i < n_1 \\
1-\frac{i - n_1}{n_2}, & n_1 \leq i < n_1 + n_2\\
0, &i \geq n_1 + n_2
\end{cases}
\end{equation*}
\end{subequations}
For our experiments, we use the annealing period of $n_1~+~n_2~=~1000$ steps with $n_1 = n_2 = 500$.

Since there are only about 2000 training images for resolution 256x256, the mode collapse problem is severe with the vanilla versions of all models (our model and the baselines). Thus, we adopt PacGAN2 \cite{lin2018pacgan} architecture for the discriminator, which is the same modification done for all the baselines in \cite{jolicoeur-martineau2019relative}. 

We train PairGAN with Adam \cite{kingma2015adam} using one step of discriminator per generator step. We use the same setting of hyperparameters as in baseline models: learning rate $0.0002$, $\beta_1=0.5$, $\beta_2=0.999$. We implement PairGAN in PyTorch \cite{paszke2019pytorch}.

\begin{table*}[h]
\centering
 \caption{Extended version of Table \ref{table:result}. Comparison of minimum, maximum, mean, and standard deviation of FID calculated at steps 20k, 30k, ..., 100k on different resolutions of CAT dataset \cite{zhang2008cat}. `---' means model becomes stuck in the first few iterations. Baseline results denoted with (\tdag) were extracted from \cite{jolicoeur-martineau2019relative}, not independently run in our experiments.}
 \vskip 0.15in
 \begin{tabular}{l | c c c c } 
 \toprule
 Loss & Min & Max & Mean & SD \\
 \midrule
 \midrule
 \multicolumn{5}{c}{$64 \times 64$ images}\\
 \midrule
 \midrule
 SGAN & 13.51 $\pm$ 1.73 & 41.89 $\pm$ 25.57 & 23.78 $\pm$ 8.37 & 8.81 $\pm$ 6.68 \\
 RSGAN\tdag & 19.03 & 42.05 & 32.16 & 7.01 \\
 RaSGAN\tdag & 15.38 & 33.11 & 20.53 & 5.68 \\
 \hline
 LSGAN\tdag & 20.27 & 224.97 & 73.62 & 61.02 \\
 RaLSGAN\tdag & \textbf{11.97} & \textbf{19.29} & \textbf{15.61} & 2.55 \\
 \hline
 HingeGAN\tdag & 17.60 & 50.94 & 32.23 & 14.44 \\
 RaHingeGAN\tdag & 14.62 & 27.31 & 20.29 & 3.96 \\
 \hline
 RSGAN-GP\tdag & 16.41 & 22.34 & 18.20 & 1.82 \\
 RaSGAN-GP\tdag & 17.32 & 22 & 19.58 & \textbf{1.81} \\
 \hline
 \textbf{PairGAN (ours)} & 12.66 $\pm$ 2.24 & 20.90 $\pm$ 4.27 & 16.38 $\pm$ 2.84 & 2.23 $\pm$ 0.56 \\
 \midrule
 \midrule
 \multicolumn{5}{c}{$128 \times 128$ images}\\
 \midrule
 \midrule
 SGAN & 27.35 $\pm$ 7.92 & 57.76 $\pm$ 13.53 & 40.17 $\pm$ 8.66 & 9.34 $\pm$ 2.37 \\
 RaSGAN\tdag & 21.05 & 39.65 & 28.53 & 6.52 \\
 \hline
 LSGAN\tdag & 19.03 & 51.36 & 30.28 & 10.16 \\
 RaLSGAN\tdag & \textbf{15.85} & 40.26 & 22.36 & 7.53 \\
 \hline
 \textbf{PairGAN (ours)} & 17.30 $\pm$ 0.48 & \textbf{29.32 $\pm$ 3.00} & \textbf{21.92 $\pm$ 0.39} & \textbf{3.76 $\pm$ 0.65} \\
 \midrule
 \midrule
 \multicolumn{5}{c}{$256 \times 256$ images}\\
 \midrule
 \midrule
 SGAN & 69.64 $\pm$ 39.32 & 344.55 $\pm$ 16.83 & 208.99 $\pm$ 44.11 & 104.08 $\pm$ 18.32 \\
 RaSGAN\tdag & \textbf{32.11} & 102.76 & 56.64 & 21.03 \\
 \hline
 LSGAN\tdag & --- & --- & --- & --- \\
 RaLSGAN\tdag & 35.21 & 299.52 & 70.44 & 86.01 \\
 \hline
 WGAN-GP\tdag & 155.46 & 437.48 & 341.91 & 101.11 \\
 \hline
 \textbf{PairGAN (ours)} & 35.35 $\pm$ 1.15 & \textbf{64.77 $\pm$ 11.87} & \textbf{45.21 $\pm$ 2.59} & \textbf{9.49 $\pm$ 3.45} \\
 \bottomrule
 \end{tabular}
\label{table:app_result}
\end{table*}

\subsection{Further discussion}
\label{app:experiment_further}

Table \ref{table:app_result} provides an extended comparison with more baselines. These additional baselines are: least square GAN (LSGAN) \cite{mao2017least}, Hinge-loss GAN (HingeGAN) \cite{miyato2018spectral}, Wassertein GAN with gradient penalty (WGAN-GP) \cite{gulrajani2017improved}, Relativistic average LSGAN (RaLSGAN), Relativistic average HingeGAN (RaHingeGAN) \cite{jolicoeur-martineau2019relative}.

We also provide a comparison of FID trajectories for PairGAN and SGAN on different resolutions in Figure~\ref{fig:app_stability}. The performance of SGAN is unstable both along the trajectory and across trials. PairGAN, in comparison, is much more stable in both aspects for all resolutions.

We make the following observations by qualitatively examining the samples generated by SGAN and PairGAN at different stages of training. For 64x64, SGAN already suffers from mode collapse in 1 out of 3 runs. PairGAN does not manifest this issue in any of 3 runs. For 128x128, SGAN suffers from severe mode collapse in all three runs, whereas we have not observed this problem for PairGAN. For 256x256, SGAN is very unstable and the generator constantly rotates between generating low quality images and random noise. The quality of the samples generated by PairGAN is stable over the course of training and across the training runs.

\subsection{Examples}
\label{app:experiment_examples}

\begin{figure*}[h]
\centering
\includegraphics{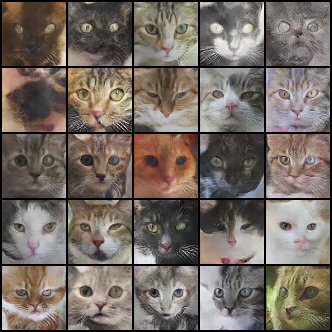}
\caption{Examples of $64\times64$ images generated with PairGAN}
\label{fig:samples_64}
\end{figure*}

\begin{figure*}[h]
\centering
\includegraphics[width=0.9\textwidth]{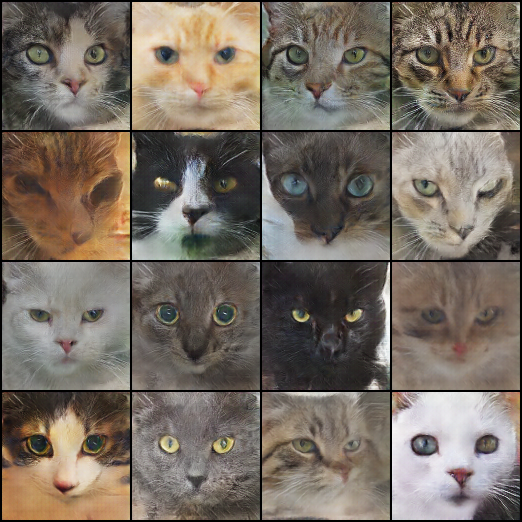}
\caption{Examples of $128\times128$ images generated with PairGAN}
\label{fig:samples_128}
\end{figure*}

\begin{figure*}[h]
\centering
\includegraphics[width=0.9\textwidth]{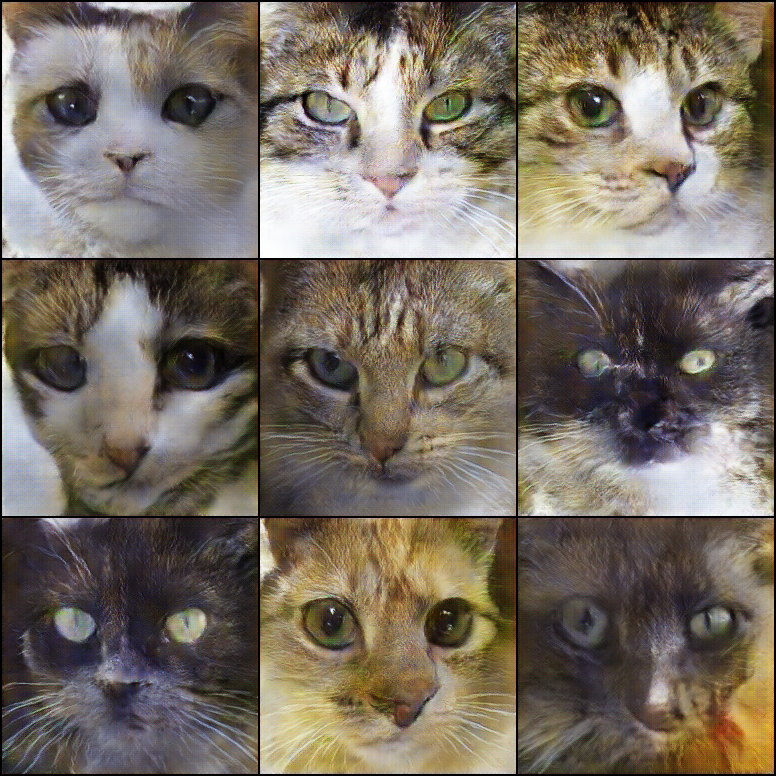}
\caption{Examples of $256\times256$ images generated with PairGAN}
\label{fig:samples_256}
\end{figure*}

Figures \ref{fig:samples_64}, \ref{fig:samples_128}, and \ref{fig:samples_256} show samples generated by PairGAN for resolutions 64x64, 128x128, and 256x256 respectively. While we resize the 128x128 and 256x256 samples in order to fit the figures in one page, we provide the original images in the code repository.

\end{document}